\documentclass{article}

\usepackage{PRIMEarxiv}
\usepackage{caption} 
\usepackage[T1]{fontenc}    
\usepackage{hyperref}       
\usepackage{url}            
\usepackage{booktabs}       
\usepackage{amsfonts}       
\usepackage{nicefrac}       
\usepackage{microtype}      
\usepackage{lipsum}
\usepackage{fancyhdr}       
\usepackage{graphicx}       
\usepackage{amsmath}
\usepackage{float}
\usepackage{subfig}
\usepackage{mathtools}
\usepackage{authblk}

\graphicspath{{media/}}     

\title{Deep Variational Lesion-Deficit Mapping}
\author[1]{Guilherme Pombo}
\author[1]{Robert Gray}
\author[1]{Amy P.K. Nelson}
\author[1]{Chris Foulon}
\author[1]{John Ashburner}
\author[1]{Parashkev Nachev}

\affil[1]{UCL Queen Square Institute of Neurology, University College London, London, UK}

\begin{document}
\maketitle

\begin{abstract}
Causal mapping of the functional organisation of the human brain requires evidence of \textit{necessity} available at adequate scale only from pathological lesions of natural origin. This demands inferential models with sufficient flexibility to capture both the observable distribution of pathological damage and the unobserved distribution of the neural substrate. Current model frameworks---both mass-univariate and multivariate---either ignore distributed lesion-deficit relations or do not model them explicitly, relying on featurization incidental to a predictive task. Here we initiate the application of deep generative neural network architectures to the task of lesion-deficit inference, formulating it as the estimation of an expressive hierarchical model of the joint lesion and deficit distributions conditioned on a latent neural substrate. We implement such deep lesion deficit inference with variational convolutional volumetric auto-encoders. We introduce a comprehensive framework for lesion-deficit model comparison, incorporating diverse candidate substrates, forms of substrate interactions, sample sizes, noise corruption, and population heterogeneity. Drawing on 5500 volume images of ischaemic stroke, we show that our model outperforms established methods by a substantial margin across all simulation scenarios, including comparatively small-scale and noisy data regimes. Our analysis justifies the widespread adoption of this approach, for which we provide an open source implementation: \url{https://github.com/guilherme-pombo/vae_lesion_deficit}

\end{abstract}


\section{Introduction}
\label{intro}

Credible inference to the functional anatomy of the brain requires evidence of the \textit{necessity} of any candidate neural substrate \cite{rorden2004using,siddiqi2022causal}. Founded on the observation of selective functional deficits under focal neural disruption, such inference relies on capturing the wide causal field \cite{mackie1980cement} of material factors the richly interconnected nature of the brain potentially brings into play. This greatly complicates the task, for two interacting reasons. First, where the disruption is non-randomly distributed across multiple locations---inevitable with natural lesions and clinical interventions---a critical region may be obscured by spurious associations owing to systematic patterns of incidental, 'collateral' damage \cite{yee_parashkev}. Second, where the substrate is non-linearly distributed across multiple regions---overwhelmingly likely in the brain \cite{faure2001there}---identifying a single critical locus may depend on the joint status of many others, some outside the contemporaneously sampled field. In both cases, complex interactions between anatomical loci---driven by pathology in the former, function in the latter---may combine to obscure and distort the inferred picture.

The problem can be illustrated in a simple synthetic system, where a hypothetical behaviour depends on three clusters located on a square grid, and `deficits' are generated by intersection with a set of ellipsoid `lesions' above a critical threshold (Figure 1). Conventional mass-univariate inference, applied voxel-wise in ignorance of correlations across regions, may be broadly relied upon to succeed in retrieving the clusters where the dependence is simple---e.g. inclusive disjunction---and the distribution of lesion features benign---e.g. uniformly oriented and located---but not where either is not. Crucially, the two distorting effects may combine to produce a highly significant inferred region very different from the target substrate (Figure 1, bottom right).

\begin{figure}[H]
    \centering
    \includegraphics[width=12cm]{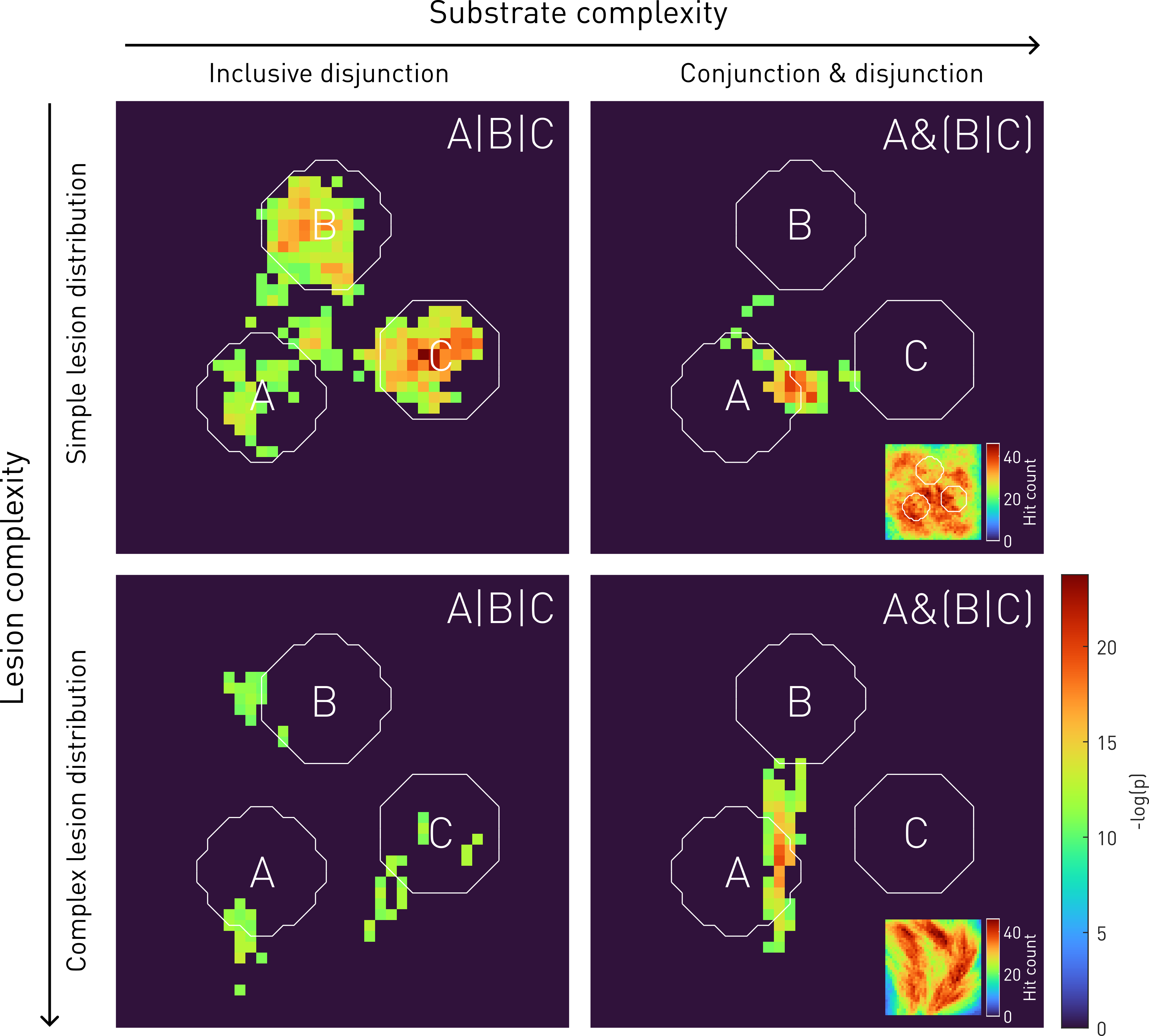}
    \caption{Demonstration with a planar $42\times42$ pixel synthetic model of the impact of substrate and lesion complexity on the fidelity of lesion-deficit inference. Across separate numerical simulations, three hypothetically critical areas---A, B, \& C---modelled as thresholded Gaussian blobs of fixed size, vary in their relation to a deficit from a simple inclusive disjunction, A|B|C, (left column) to a complex conjunction and disjunction, A\&(B|C) (right column), and are mapped with lesions drawn from a simple lesion distribution, with uniformly distributed parameters, (top row) or a complex lesion distribution, with spatially structured parameters, (bottom row). For each condition, a deficit generating process based on 10\% intersection of a lesion with the substrate as defined above is used to create a synthetic ground truth from 400 uniformly randomly distributed single synthetic lesions in the form of binary ellipses with a major to minor diameter ratio uniformly randomly varying between 0.1 and 0.4, yielding a deficit label vector. The orientation of the lesions in the simple lesion condition is uniformly random, and in the complex lesion condition varies systematically with location, yielding pixel-wise 'hit' maps of comparable density (inset plots). Lesion-deficit inference is performed with Fisher’s exact test applied in a pixel-wise manner to generate a field of asymptotic p-values for each condition, shown as the negative log thresholded at a Bonferroni-corrected $\alpha$ of 0.05. Note the interaction between lesion and substrate complexity in driving mislocalization where the inferential method ignores correlations across loci.}
    \label{fig:ldm_diagram}
\end{figure}

If the potential for substantial error with simple inferential methods is plain, estimating bounds on its magnitude is very difficult. First, the core task has no ground truth: the critical neural substrate is what we \textit{infer}. Though prediction of deficits conditional on a lesion may be a rough guide, it addresses a different question---the optimal location of the deficit discriminant boundary---and extraneous, non-anatomical factors may set an unquantifiably low ceiling on maximum achievable fidelity \cite{tianbo_ldm}. Equally, consensus with non-disruptive methods such as functional imaging cannot be decisive, for in disagreement the disruptive method would be favoured owing to its greater inferential power \cite{pustina}. Second, the connective organization of the brain provides only minimal constraint on the space of possible functional topologies, and the morphological structure of the most abundant source of disruptive evidence---ischaemic lesions---exhibits a distribution only a highly expressive model could conceivably learn with good fidelity. This leaves the joint system of substrate-lesion interactions analytically intractable.

The validation of any candidate inferential approach therefore inevitably requires extensive numerical simulations of semi-synthetic ground truths \cite{bates_vlsm, zhang_svr, svr_lsm_2, svrlsm_eval}, and provides assurance only in proportion to the range and density of lesion-deficit scenarios surveyed. We can, however, prescribe the essential features a good model architecture must posses, for the problem formulation dictates them. First, it must be multivariate, for the only defence against the distorting effects of interactions between loci is to model them \cite{pustina, zhang_svr}. Second, it must be non-parametric, for no assumptions about the underlying distributions can be confidently made. Third, it must be generative, for our primary objective is to estimate the distribution of the critical neural substrate. Fourth, it must be highly expressive, for neither substrates nor lesions can be assumed to be simple in their structure. Fifth, it must be robust to noise and omissions, for the experimental context makes both unavoidable. Finally, it must be statistically efficient, for data jointly on function and disruption in the human brain is rarely available at scale.

No current approach satisfies all these requirements: here we set out to create one, drawing on Bayesian deep generative models of volumetric data \cite{pixelcnn, nick_encoder, diffeo_gan}. Our principled solution, Deep Lesion Mapping (DLM), employs a variational auto-encoder \cite{vae} to learn the joint distribution of lesions and deficit labels in terms of an anatomical `latent' neural substrate. Using 8 Nvidia V100 GPUs over roughly $12000$ hours of compute, we subject our method to the most extensive validation of lesion-deficit mapping ever conducted, ranging across diverse candidate substrates, forms of substrate interactions, sample sizes, noise corruption, and population heterogeneity, with the largest known collection of registered images of acute ischaemic stroke: 5500 volumes derived from 4788 unique patients. We compare DLM with mass-univariate inference, and the two multivariate methods most commonly cited in the field, establishing a new state-of-the-art level of performance by a substantial margin across all modelled scenarios. We provide the theoretical underpinnings of the success of the approach, and outline a semi-supervised extension.

\begin{figure}[H]
    \centering
    \includegraphics[width=12cm]{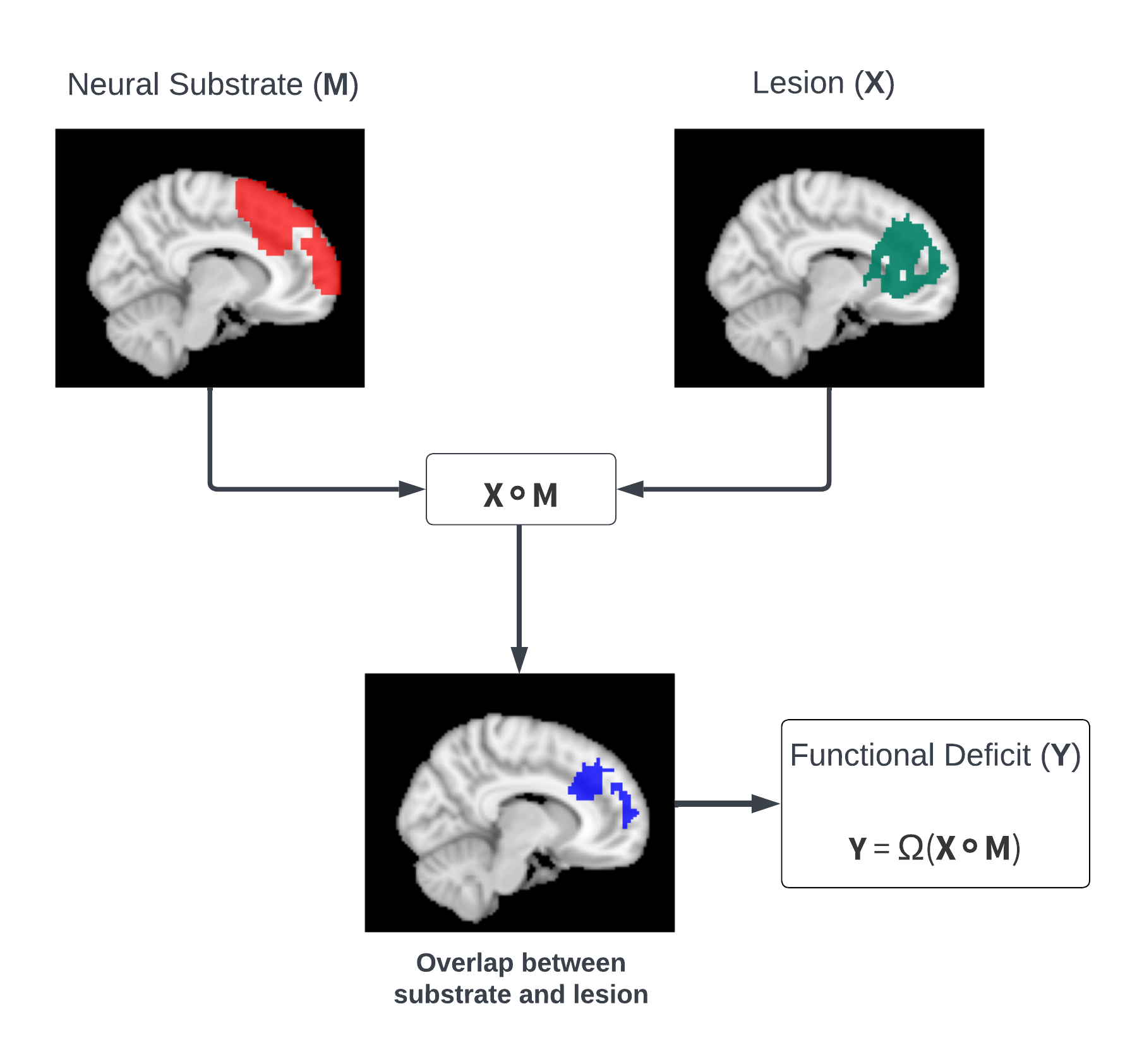}
    \caption[Diagrammatic explanation of LDM assumptions]{A diagrammatic explanation of the assumptions made in LDM.  The assumptions have been used extensively in literature \cite{bates_vlsm, zhang_svr, pustina, svrlsm_eval} to carry out simulations, given a simulated neural substrate, $\mathbf{M}$ and real lesions, $\mathbf{X}$, one can simulate the functional deficits, $\mathbf{Y}$ as a function of the overlap of the lesion and the neural substrate.}
    \label{fig:ldm_diagram}
\end{figure}

\section{Related Work}

\subsection{Lesion-deficit mapping}
\label{ldm_intro}

Lesion-deficit mapping (LDM) relies on the assumption that when neural substrates, $\mathbf{M}$, are damaged by lesions, $\mathbf{X}$, observable functional deficits, $\mathbf{Y}$, result. Moreover, $\mathbf{Y}$ is related to $\mathbf{M}$ and $\mathbf{X}$ by some function, $\Omega$, of the spatial overlap between substrates and lesions:

\begin{equation}
    \mathbf{Y} = \Omega(\mathbf{M} \circ \mathbf{X})
\end{equation}

On these assumptions, illustrated in Figure \ref{fig:ldm_diagram}, rests the inference of neural substrates that have not been, and cannot possibly be, observed directly.

Inferring the true neural substrate is the goal of LDM and the inferred neural substrate is called either the \textbf{lesion-deficit map} or \textbf{lesion-symptom map}. Since the true neural substrates are unknown, LDM methods are often evaluated by simulating deficit scores with a relatively simple function $\Omega$ and simulated neural substrate $\mathbf{M}$ \cite{bates_vlsm, zhang_svr, pustina}. Many such simulations are described in \S \ref{functional_neural substrate_section} and \ref{score_simulation}. \newline

\subsection{Voxel-based lesion–symptom mapping}
\label{VLSM}

The most widely used method in the lesion-deficit literature is Voxel-based Lesion–Symptom Mapping (VLSM) \cite{bates_vlsm}, an adaptation of the voxel-wise, mass-univariate inference pioneered by Statistical Parametric Mapping (SPM, \cite{friston2003statistical}). Here the relationship between each voxel and a deficit (our preferred term given that a deficit may be asymptomatic) is established by multiple statistical tests independently performed at each voxel. Where the deficit labels are binomial, either a Fisher or a Liebermeister test is used \cite{improving_lsm}; where they are continuous, either  two-sample t-tests \cite{t_test} or the Brunner-Munzel (BM) test \cite{bm_test}. The BM test is often preferred to the t-test, because it does not require the assumption of equal variances between groups \cite{pustina, improving_lsm}. To avoid dependence on specific assumptions, here we always evaluate both BM and Fischer tests and present the best performing results, presenting VLSM in the best possible light.\newline

The simplicity and computational efficiency of VLSM explain its wide use. As we have seen, however, a mass-univariate framework cannot account for complex correlations across loci\cite{flaws_vlsm, pustina, univariate_shortcomings_1, univariate_shortcomings_2}, and will always be vulnerable to distortion. This has been repeatedly shown to result in substantial mislocalization that cannot be rectified by larger data scales, indeed is merely entrenched by them \cite{yee_parashkev}. These failure cases and others \cite{parashkev_brain, pustina, flaws_vlsm} establish the need for multivariate modelling. \newline

\subsection{Lesion–symptom mapping with multivariate sparse canonical correlations (SCCAN)}
\label{SCCAN}

Canonical correlation analysis (CCA) \cite{cca} is a multivariate method that given a paired dataset, $\mathbf{x} \in \mathbb{R}^{n \times p}$ and $\mathcal{\mathbf{y}} \in \mathbb{R}^{n \times q}$, computes two linear projections, $\mathbf{w}_x$ and $\mathbf{w}_y$, such that the projected data for each modality is maximally linearly correlated. More formally, it has the following objective:

\begin{equation}
    \underset{\mathbf{w}_{x}, \mathbf{w}_{y}}{\text{argmax}} \operatorname{corr}\left(\mathbf{x} \mathbf{w}_{x}, \mathbf{y} \mathbf{w}_{y}\right) = \underset{\mathbf{w}_{x}, \mathbf{w}_{y}}{\text{argmax}} \frac{\mathbf{w}_{x}^{\top} \boldsymbol{\Sigma}_{x y} \mathbf{w}_{y}}{\sqrt{\mathbf{w}_{x}^{\top} \boldsymbol{\Sigma}_{x x} \mathbf{w}_{x}} \sqrt{\mathbf{w}_{y}^{\top} \boldsymbol{\Sigma}_{y y} \mathbf{w}_{y}}}.
\end{equation}

Here, $\boldsymbol{\Sigma}_{x y}$ is the covariance matrix between $\mathbf{x}$ and $\mathbf{y}$. One can introduce sparsity and positivity constraints into the optimisation objective, $0 < \left\|\mathbf{w}_{x}\right\|^{2} \leq K, 0< \left\|\mathbf{w}_{y}\right\|^{2} \leq K$, where $K \leq 1$, through penalised matrix decomposition algorithms \cite{witten_cca}. These constraints are often used in neuroimaging to improve interpretability by making the resulting projections more focused on important regions of interest. The authors of \cite{pustina} show that when $\mathbf{x}$ is a set of stroke lesions and $\mathbf{y}$ a set of associated deficits, the CCA produced projections $\mathbf{w}_{x}$ can be used as a lesion-deficit maps and that these maps are more accurate than the ones produced by VLSM. \newline

In order to avoid the use of permutation thresholding methods, the authors of \cite{pustina} grid-search across sparsity constraints $K$ and keep the lesion-deficit maps which can most accurately recover the behavioural scores on a validation set. The projection which obtains the most accurate behavioural scores on the cross validation is then used as the final inferred lesion-deficit map. They call their method \textbf{SCCAN}.

\subsection{Multivariate lesion-symptom mapping using support vector regression}
\label{SVR}
 
In \cite{zhang_svr} the authors propose formulating lesion-symptom mapping as a regression problem and solving it using Support Vector Regression (SVR) \cite{svr}, to introduce increased modelling capacity in the form of a multivariate non-linear solution. Given a function (kernel), $\boldsymbol{\phi}$, the regression problem is then posed as:

\begin{equation}
    \mathbf{y} = \mathbf{W}^{T} \boldsymbol{\phi}(\mathbf{x})  + b
\end{equation}

where $\mathbf{W}$ is a matrix with the fitting coefficients and $b$ a scalar representing the fitting errors. If $\boldsymbol{\phi}$ is nonlinear, then in order to project the feature space in which $\boldsymbol{\phi}(\mathbf{x})$ operates onto the image space one requires inverse transformations, which generally cannot be found analytically. Methods exist to approximate these transformations \cite{preimage_1, preimage_2} but they require computationally expensive and unstable optimisations and therefore the authors of \cite{zhang_svr} derive their own approximate back-projection which is cheap to compute, but less precise than traditional methods and unquantified in terms of spatial bias \cite{svrlsm_eval}. \newline

To perform statistical inference with the SVR-LSM model, the authors use Monte Carlo permutation testing of the model weights, citing the p-value at each voxel. Note that this p-value does not test the null hypothesis that a voxel is not associated with the behaviour, but rather the null that the weights of the model are randomly distributed. It is therefore merely an indicator of feature relevance within a model. 
\newline

\subsection{Thresholding in existing LDM methods}
\label{thresholding_section}

Most established methods for inferring lesion-deficit maps are frequentist \cite{bates_vlsm, zhang_svr, yee_parashkev}. Here inferring the necessity of the substrate rests on rejecting the null hypothesis that a deficit does not depend on a particular region of the brain. \newline

One frequentist approach to deriving an inferential criterion is to resample the observed data repeatedly, assigning different outcome values to each observation from among the set of actually observed outcomes \cite{bates_vlsm, zhang_svr}. Such \textit{permutation tests} and can be understood as surrogate data testing, where the surrogate data under the null hypothesis are obtained through permutations of the original data. In practice, there are too many possible orderings of the data and hence practitioners often employ asymptotically equivalent Monte Carlo subsamples  the total possible permutations.\cite{random_perm_tests}. \newline

Where multiple comparisons are performed---inevitable with mass-univariate approaches---correction for the baseline error rate is required. There are two families of multiple comparison correction methods:
\begin{enumerate}
    \item \textbf{Familywise Error Rate} (FWER): attempts to control the probability of finding any false positives.
    \item \textbf{False Discovery Rate} (FDR): attempts to control the proportion of false positives among rejected tests
\end{enumerate}

Each method has its respective merits and demerits, however, FWER methods have been shown to be more robust in the context of brain imaging \cite{fwer_robust} and so we use them for multiple comparison correction throughout \S \ref{inference_experiments}. \newline

Such correction is necessary only where multiple comparisons are performed, and spatial inference rests on the associated statistics \cite{pustina}. Where the inference rests on validation and test sets, it is not required. \newline

\section{A latent substrate model of lesions and deficits}
\label{vae_ldm_theory}

Our approach is to model the joint distribution of lesions and deficits, $P(\mathbf{X}, \mathbf{Y})$, in terms of a `latent' neural substrate. We use variational inference \cite{bishop2006pattern} to bound the likelihood in terms of this latent variable. We implement and optimise this bound using a variational auto-encoder (VAE) \cite{vae}. A VAE is a latent variable model that is capable of capturing even complex distributions \cite{vdvae}. The latent neural substrate interacts anatomically with the lesion according to a biologically plausible deficit-generating process---intersection---determining how lesions convert into deficits. As the model learns, we thus infer the anatomical basis for the underlying lesion-deficit relationships. \newline

Using model parameters, $\boldsymbol{\theta}, \boldsymbol{\psi}, \boldsymbol{\phi}$ and a latent variable, $\mathbf{Z}$, we express the marginal likelihood in terms of a prior, a likelihood, and an approximate posterior:

\begin{equation}
\begin{split}
    P_{\boldsymbol{\theta}}(\mathbf{X}, \mathbf{Y}) & = \int_\mathbf{z} P_{\boldsymbol{\theta}}(\mathbf{X}, \mathbf{Y}| \mathbf{Z}=\mathbf{z}) P_{\boldsymbol{\psi}}(\mathbf{Z}=\mathbf{z}) d\mathbf{z} \\
    & = \int_\mathbf{z} P_{\boldsymbol{\theta}}(\mathbf{X}, \mathbf{Y}| \mathbf{Z}=\mathbf{z}) P_{\boldsymbol{\psi}}(\mathbf{Z}=\mathbf{z}) \frac{Q_{\boldsymbol{\phi}}(\mathbf{Z}=\mathbf{z}|\mathbf{X}, \mathbf{Y})}{Q_{\boldsymbol{\phi}}(\mathbf{Z}=\mathbf{z}|\mathbf{X}, \mathbf{Y})} d\mathbf{z} \\
    & = \mathbb{E}_{\mathbf{z}\sim Q_{\boldsymbol{\phi}}(\mathbf{Z}|\mathbf{X}, \mathbf{Y})} \frac{P_{\boldsymbol{\theta}}(\mathbf{X}, \mathbf{Y}| \mathbf{Z}=\mathbf{z}) P_{\boldsymbol{\psi}}(\mathbf{Z}=\mathbf{z})}{Q_{\boldsymbol{\phi}}(\mathbf{Z}=\mathbf{z}|\mathbf{X}, \mathbf{Y})}.
\end{split}
\end{equation}

To do so we parameterise the distribution with:
\begin{itemize}
    \item $\boldsymbol{\theta}$ for the joint likelihood. In the context of a VAE this corresponds to the Decoder.
    \item $\boldsymbol{\phi}$ for the posterior distribution. In the context of a VAE this corresponds to the Encoder.
    \item For the prior distribution we simply use the unit Normal distribution $\mathcal{N}(\mathbf{0}, \mathbf{I})$ and denote it $\boldsymbol{\psi}$.
\end{itemize}

Applying natural logarithms and using Jensen's inequality yields the variational lower bound:

\begin{equation}
\begin{split}
   \log P_{\boldsymbol{\theta}}(\mathbf{X}, \mathbf{Y}) & \geq \mathbb{E}_{\mathbf{z}\sim Q_{\boldsymbol{\phi}}} [\log P_{\boldsymbol{\theta}}(\mathbf{X}, \mathbf{Y}| \mathbf{Z}=\mathbf{z}) \\
   & \hphantom{=} + \log P_{\boldsymbol{\psi}}(\mathbf{Z}=\mathbf{z}) - \log Q_{\boldsymbol{\phi}}(\mathbf{Z}=\mathbf{z}|\mathbf{X}, \mathbf{Y}) ] \\
& = \mathbb{E}_{\mathbf{z}\sim Q_{\boldsymbol{\phi}}} [\log P_{\boldsymbol{\theta}}(\mathbf{X}, \mathbf{Y}| \mathbf{Z}=\mathbf{z}) ] \\
& \hphantom{=} + \mathbb{E}_{\mathbf{z}\sim Q_{\boldsymbol{\phi}}} [\log P_{\boldsymbol{\psi}}(\mathbf{Z}=\mathbf{z}) - \log Q_{\boldsymbol{\phi}}(\mathbf{Z}=\mathbf{z}|\mathbf{X}, \mathbf{Y})] \\
& = \mathbb{E}_{\mathbf{z}\sim Q_{\boldsymbol{\phi}}} [\log P_{\boldsymbol{\theta}}(\mathbf{X}, \mathbf{Y}| \mathbf{Z}=\mathbf{z})] \\
& \hphantom{=} - D_{KL}(Q_{\boldsymbol{\phi}}(\mathbf{Z}|\mathbf{X}, \mathbf{Y}) || P_{\boldsymbol{\psi}}(\mathbf{Z})),
\end{split}
\end{equation}

where $D_{KL}$ is the KL divergence. Given the assumptions of lesion-deficit mapping discussed in \S \ref{ldm_intro}, we want the deficits, $\mathbf{Y}$, to be dependent on the lesions, $\mathbf{X}$, and the neural substrates, $\mathbf{M}$. Therefore, we use two parameters $\boldsymbol{\theta}$ and $\boldsymbol{\gamma}$, as well as the definition of conditional probability to factorise the likelihood as follows:

\begin{equation}
   \log P_{\boldsymbol{\theta}}(\mathbf{X}, \mathbf{Y}| \mathbf{Z}) = \log P_{\boldsymbol{\gamma}}(\mathbf{Y}|\mathbf{X},\mathbf{Z}) + \log P_{\boldsymbol{\theta}}(\mathbf{X}|\mathbf{Z}).
\end{equation}

The lesions, $\mathbf{X}$, are conventionally treated as binary masks, and so we model them using a Bernoulli random variable. Given binary functional deficits, we compute, $P_{\boldsymbol{\gamma}}(\mathbf{Y}|\mathbf{X},\mathbf{Z})$ using a simple linear relationship in which $\boldsymbol{\gamma}(\mathbf{z})$ plays the role of the neural substrate $\mathbf{M}$ in \S \ref{ldm_intro}:

\begin{equation}
    P_{\boldsymbol{\gamma}}(\mathbf{Y}|\mathbf{X}=\mathbf{x},\mathbf{Z}=\mathbf{z}) = \text{Bernoulli} (\sigma(\mathbf{x}^{T} \boldsymbol{\gamma}(\mathbf{z})))
\end{equation}

When modelling continuous deficits, we use two sets of parameters $\boldsymbol{\gamma}_{\mu}$ and $\boldsymbol{\gamma}_{\sigma}$, to model the mean and standard deviation of a Normal distribution:

\begin{equation}
    P_{\boldsymbol{\gamma}}(\mathbf{Y}=\mathbf{y}|\mathbf{X}=\mathbf{x},\mathbf{Z}=\mathbf{z}) = \mathcal{N} (\mathbf{x}^{T} \boldsymbol{\gamma}_{\mu}(\mathbf{z})), \mathbf{x}^{T} \boldsymbol{\gamma}_{\sigma}(\mathbf{z}))).
\end{equation}

During training we clamp the values of $\boldsymbol{\gamma}_{\sigma}$ to ensure they are always larger than zero. The final objective that our model optimises is:

\begin{equation}
   \mathbb{E}_{\mathbf{z}\sim Q_{\boldsymbol{\phi}}} [\log P_{\boldsymbol{\gamma}}(\mathbf{Y}|\mathbf{X},\mathbf{Z}) + \log P_{\boldsymbol{\theta}}(\mathbf{X}|\mathbf{Z})] - D_{KL}(Q_{\boldsymbol{\phi}}(\mathbf{Z}|\mathbf{X}, \mathbf{Y}) || P_{\boldsymbol{\psi}}(\mathbf{Z}))
\end{equation}

\subsection{Network architecture}
\label{network_architecture}

In the interests of promoting statistical efficiency, we choose a simple 3D convolutional implementation of our framework. CNNs have fewer parameters than fully connected networks and hence, given the sparsity of both the input data and the labels, we find them a good fit for the problem. However, as described in \cite{coord_conv}, convolutions can struggle when faced with modelling tasks that require mapping between coordinates in $(x, y, z)$ Cartesian space and coordinates in one-hot pixel space and hence we adopt their solution to the issue -- an extension of the standard convolution, in which the lesion images are concatenated with 3 additional channels containing a 3D coordinate grid. \newline

Our framework has three main components, an Encoder ($\boldsymbol{\phi}$), two Decoders ($\boldsymbol{\gamma}$ and $\boldsymbol{\theta}$) and a latent space defined by a mean $\boldsymbol{\mu}$ and a scale $\boldsymbol{\sigma}$. Both Encoders and Decoders use the same  convolutional layer setup:
\begin{enumerate}
    \item A 3D convolution with kernel size $3\times3\times3$ and stride of 1. After applying the convolution, zero-padding of $1\times1\times1$ dimensions is added to keep the shape unchanged.
    \item A 3D batch normalisation layer \cite{batch_norm} is subsequently applied to the output.
    \item A non-linearity is applied. The type of non-linearity is not critical to performance, however, we find the best performance is obtained with Gaussian error linear units \cite{gelu}.
    \item In each Encoder layer, voxels are averaged with a $2\times2\times2$ kernel with a stride of 2. In each Decoder layer, voxels are upsampled with a $2\times2\times2$ kernel and nearest neighbour interpolation. 
\end{enumerate}

For the Encoder we double the number of output channels in our convolutions every time we halve the input resolution. For the Decoders we halve the number of channels in the convolutions every time we double the output resolution. To make our model more parameter-efficient the Encoder is fully-convolutional: its final feature map has $1\times1\times1$ spatial resolution. We then apply two fully-connected layers to obtain the mean $\boldsymbol{\mu}$ and scale $\boldsymbol{\sigma}$, from which we then sample our latent variables $\mathbf{z} \sim \mathcal{N}(\boldsymbol{\mu}, \boldsymbol{\sigma})$. \newline

\begin{figure}[H]
\hspace*{-1cm}
    \centering
    \includegraphics[width=15cm]{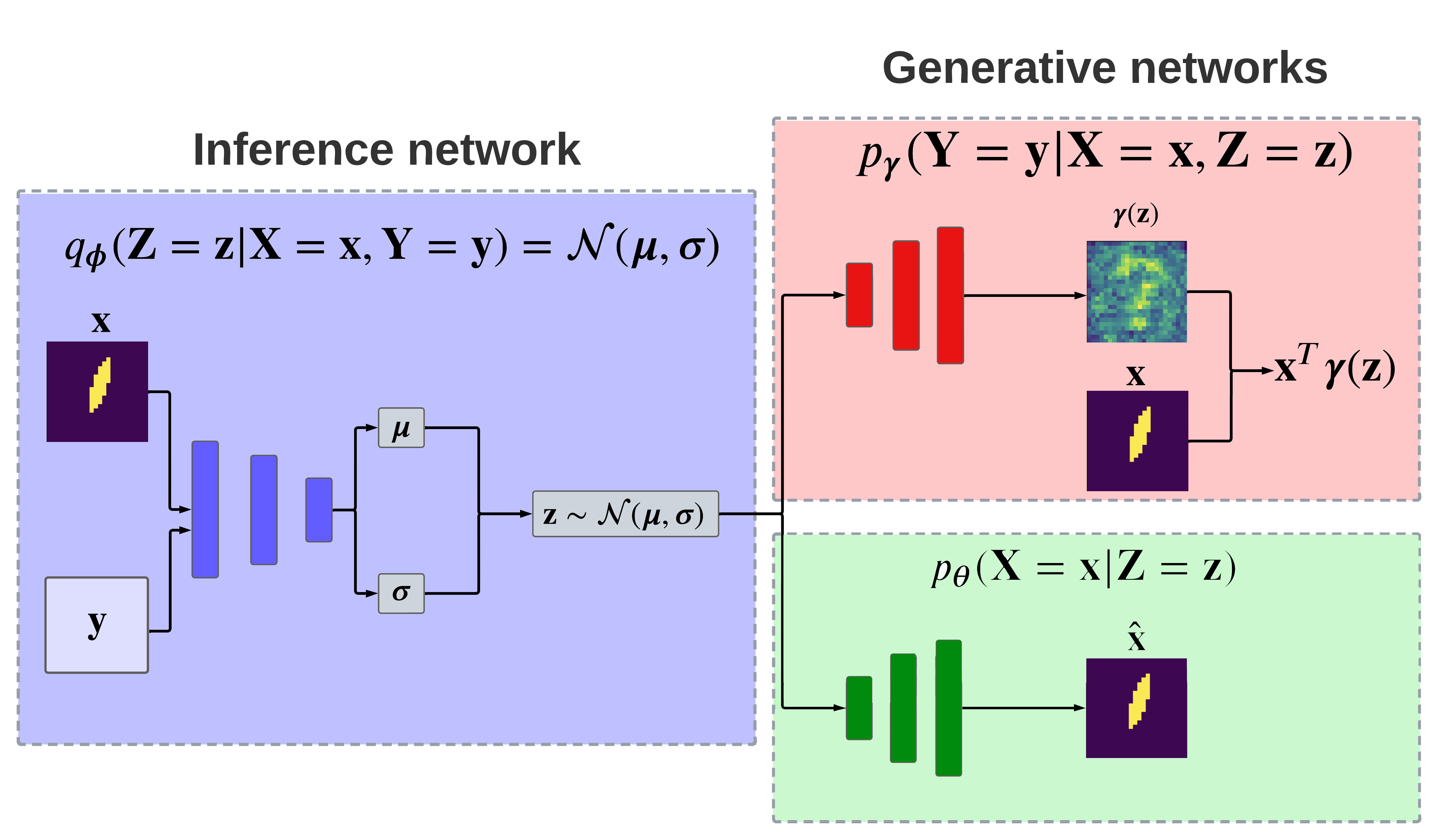}
    \caption[DLM framework architecture]{Diagram representing the DLM framework.}
    \label{fig:vae_ldm_diagram}
\end{figure}

All models are trained with early stopping using the validation set, the criterion being 20 successive epochs without a drop in validation $\log P_{\boldsymbol{\gamma}}(\mathbf{Y}|\mathbf{X},\mathbf{Z})$. The optimizer used was Adam with default parameters \cite{adam}. $L_2$ regularisation is applied to all model parameters with weight $1e-4$. A schematic of the general framework is presented in Figure \ref{fig:vae_ldm_diagram}.

\subsection{Thresholding the inferred neural substrate}

We compute the inferred neural substrate map $\hat{\mathbf{m}}$ as the expectation $\hat{\mathbf{m}} = \mathbb{E}[\mathbf{M}]$. We approximate this expectation with a finite average using $N$ samples from $P(\mathbf{Z})$:

\begin{equation}
   \mathbf{z}_{i} \sim \mathcal{N} (\mathbf{0}, \mathbf{I}) \quad \text{and} \quad
   \hat{\mathbf{m}} = \frac{1}{N} \sum_{i=1}^{N} \boldsymbol{\gamma} (\mathbf{z}_i)
\end{equation}

To generate a binary map from $\hat{\mathbf{m}}$, we use a calibration set. We search across all possible thresholds $t \in (0, 1)$, and threshold the mask using the quantile function $Q(t, \hat{\mathbf{m}})$, which for each voxel outputs $0$ if the voxel value is under the t-quantile of the mask and $1$ otherwise. We then threshold $\hat{\mathbf{m}}$, using the $t$ for which $\log P_{\boldsymbol{\gamma}}(\mathbf{Y}|\mathbf{X},\mathbf{Z})$ attains the lowest value on the calibration set. \newline

\section{Implementation of baseline methods }
\label{svm_setup}

\textbf{VLSM}: When using the Brunner- Munzel (BM) test \cite{bm_test} we use the implementation provided by  MRIcron \cite{improving_lsm} and when using Fischer's exact test we use the SciPy implementation \cite{scipy}. To calculate the FWER threshold, we computed 10,000 Monte Carlo permutation tests. After creating a full distribution of all the peak scores, the FWER threshold was set at the 95th percentile of the distribution. \newline

\textbf{SVR-LSM}: We benchmark this model using libSVM \cite{lib_svm}, and the toolbox provided in \cite{demarco} which implements an epsilon-SVR with radial basis function kernel. Similarly to the origibal SVR-LSM paper, for each simulation we grid search over SVR-LSM hyperparameters through 5-fold cross validation and choose the $C$ and $\gamma$ values which produce the best cross validation predictive performances. We grid search $C \in [1, 50]$ in increments of 5 and $\gamma \in [0.5, 10]$ in increments of 5 as well, resulting in 25 possible combinations of hyperparameters. Using those hyperparameters, the voxel-wise statistical significance levels are determined by permutation testing as described in \S \ref{thresholding_section} by comparison of pseudo-behaviour maps and the map obtained from real behavioural data. In our study we use 10,000 Monte Carlo permutation tests. In \cite{zhang_svr, demarco} it was shown that a control for the effect of lesion size is required in SVR-LSM and hence all lesion images are normalised to have a unit norm, which serves as a direct total lesion volume control (dTLVC). The final thresholding of the significance maps is done with FWER inside the toolbox provided by \cite{demarco}. \newline

\textbf{SCCAN}: We use the RStudio implementation of SCCAN provided by \cite{pustina}. For every model we run 5 fold cross-validations (CV) for L1 regularisation values in range $[0.01, 0.9]$. For each sparseness value, a model is fitted to 90\% of the subjects and the behavioral scores are predicted for the remaining 10\%. The model that achieves the highest cross validation correlation between predicted behavior and true behavior is then trained with all the data available, as done in \cite{pustina}. The map obtained with the optimal L1 value is then used as the final inferred lesion-deficit map. \newline

\section{Experiments}
\label{inference_experiments}

\subsection{Lesion data}

A set of acute ischaemic lesion masks derived from clinical imaging of patients presenting to University College London Hospitals NHS Trust with confirmed stroke was used. The total of 5500 volumes are derived from diffusion weighted imaging of 4788 unique patients. Normalised to standard space of $91 \times 109 \times 91$ voxels of 2 $\text{mm}^3$ using non-linear registration. The lesion was segmented from each image by a custom U-Net transformer model (UNETR) \cite{unetr, giles_prescriptive_inference}, and these segmentations were then manually corrected by a neurologist experienced in the task. The output of the segmentation step for each patient was a binary image, indexing the presence or absence of ischaemic stroke damage to any part of the brain. Given the number of experiments to be evaluated, unless otherwise mentioned, we resample the stroke lesions to $3\text{mm}^3$ resolution for all experiments in \S \ref{inference_experiments} to reduce time constraints. \newline

\begin{figure}[H]
    \centering
    \includegraphics[width=13cm]{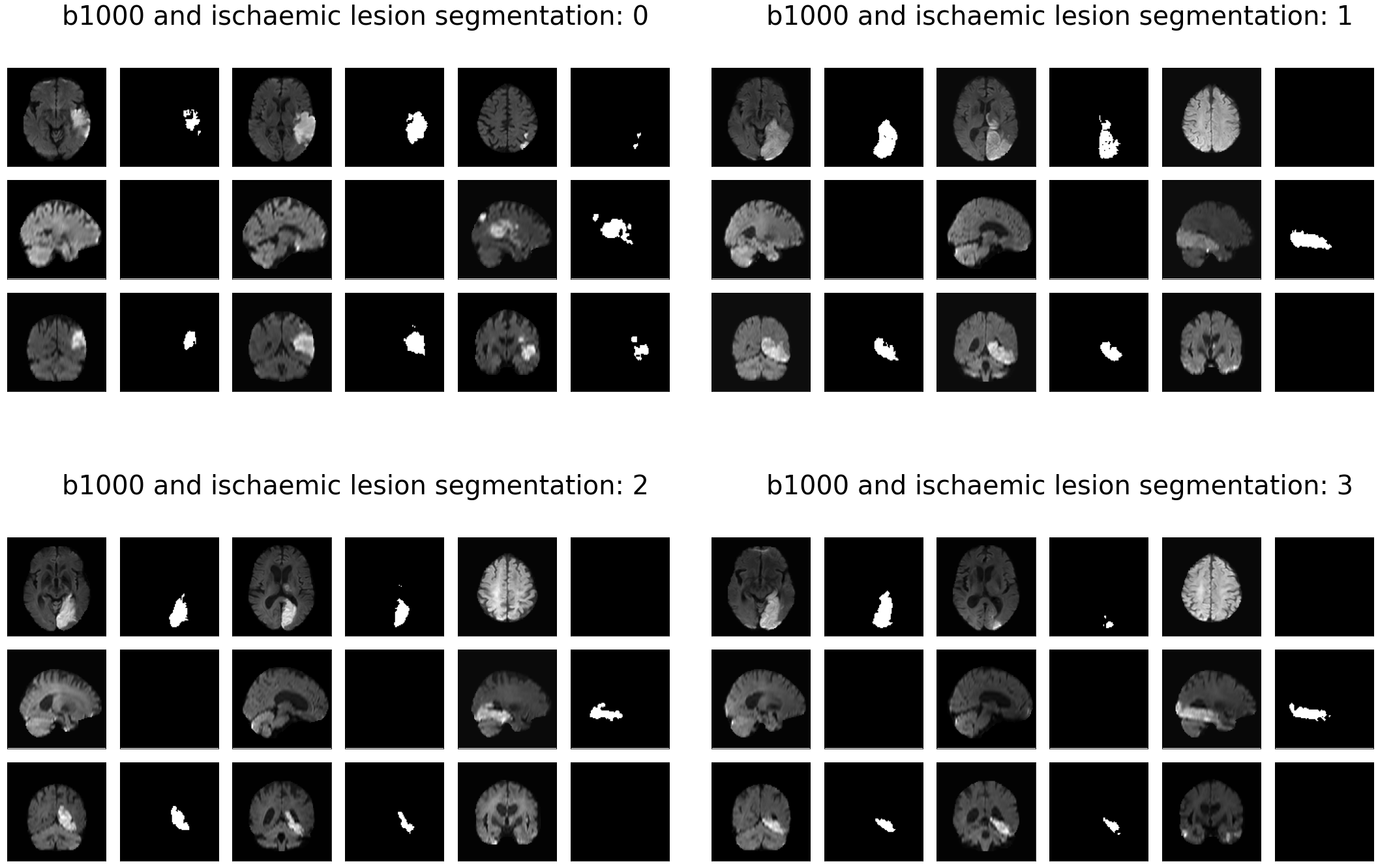}
    \caption[Examples of ischaemic brain DWIs and corresponding lesion segmentations]{Examples of b1000 scans of ischaemic brains and associated lesion segmentations.}
    \label{fig:b_1000_examples}
\end{figure}

\begin{figure}[H]
    \centering
    \includegraphics[width=12cm]{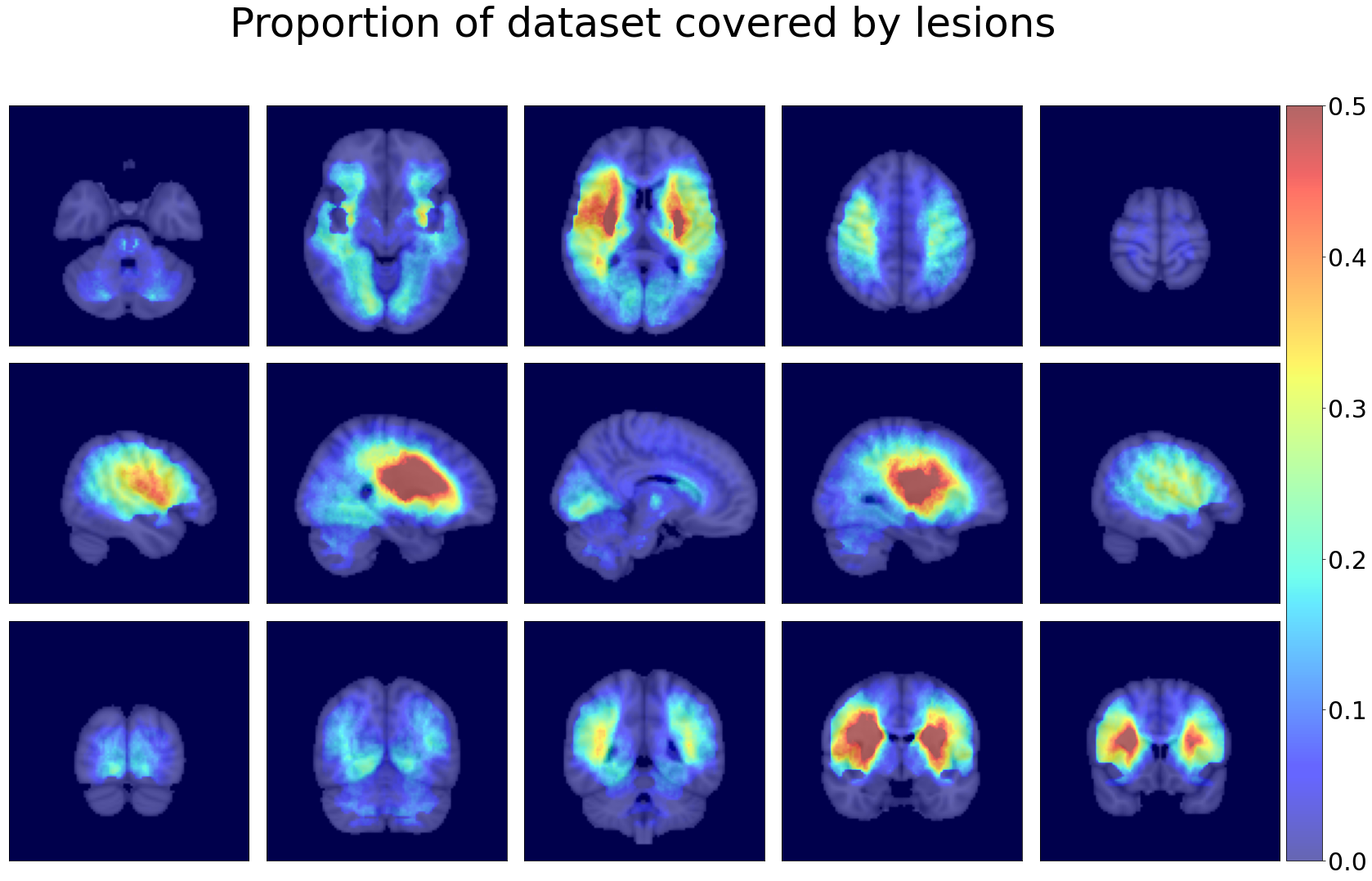}
    \caption[UCLH dataset ischaemic lesion anatomical distribution]{The proportion of the dataset covered by ischaemic stroke lesions. Clamped at 50\% for illustrative purposes}
    \label{fig:lesion_proportion}
\end{figure}

We present some example brains and their associated lesion segmentations in Figure \ref{fig:b_1000_examples} and the proportion of the dataset covered by the ischaemic lesions in Figure \ref{fig:lesion_proportion}. From looking at the lesion distributions in Figure \ref{fig:lesion_proportion} it is quite apparent that the lesion distribution is heavily biased towards certain regions and that this a factor to be considered whenever inferential models are trained. \newline

\subsection{Simulating neural substrates}
\label{functional_neural substrate_section}

Our task does not, as we have said, provide a ground truth. We must therefore simulate one \cite{yee_parashkev, zhang_svr, pustina}. This begins with defining a set of plausible, functionally-defined neural substrates. Rather than rely on structural parcellations, whose relationship to function is speculative \cite{brodmann, glasser}, we use meta-analytic functional imaging data to derive a whole-brain functional parcellation with greater plausibility, following the method described in \cite{giles_prescriptive_inference}\newline

\begin{figure}[H]
    \centering
    \includegraphics[width=12cm]{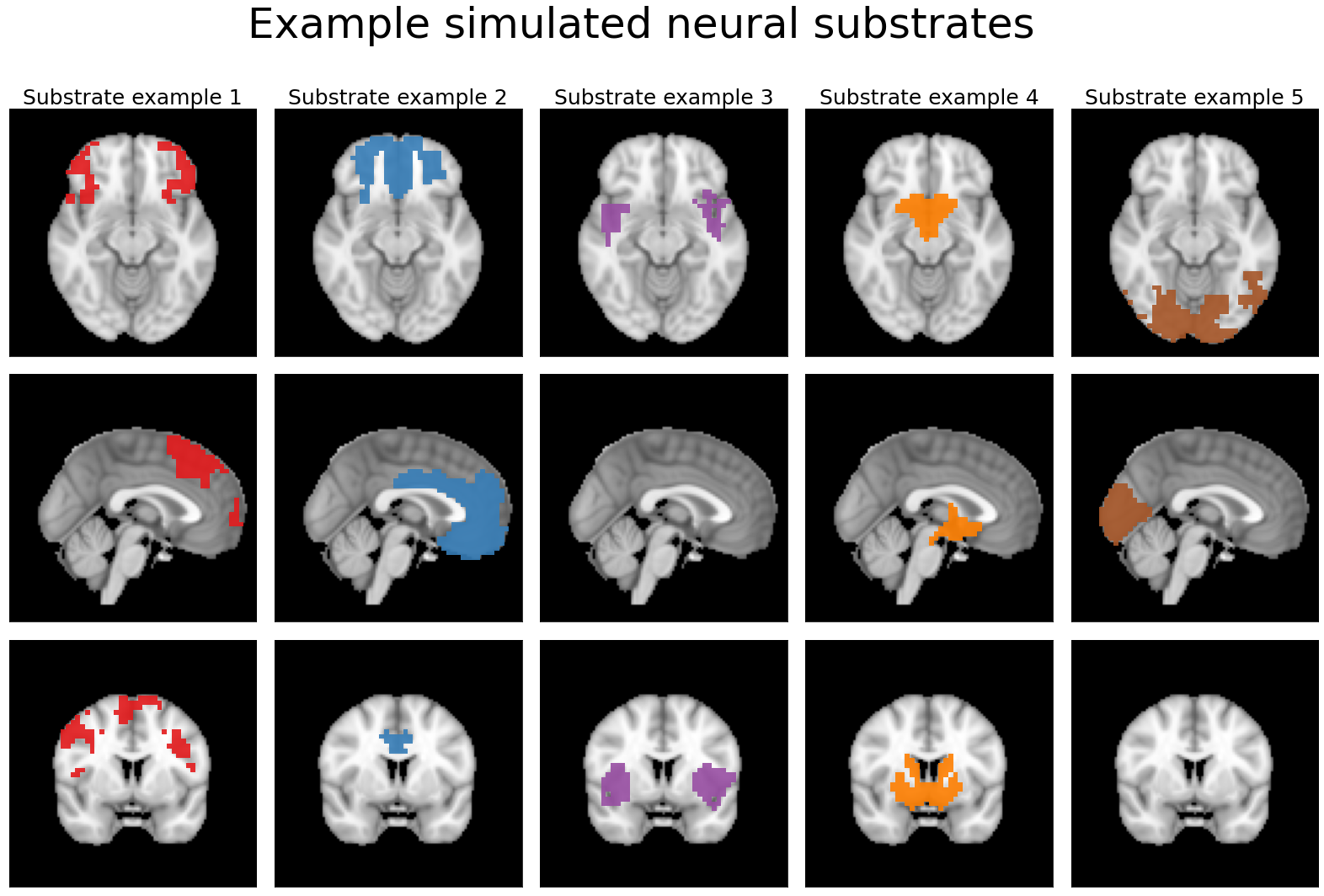}
    \caption[Example neural substrate simulations]{5 examples of the simulated neural substrates}
    \label{fig:neural_susbtrate_fig}
\end{figure}

In brief, this parcellation was derived from the NeuroQuery set \cite{neuroquery} of statistical activation maps associated with unique terms from neurological/neuroscientific corpus of literature, processed using an automated meta-analytic pipeline. From the 6134 spatial activation maps registered to MNI space, 2095 were manually selected to include only those corresponding to terms describing cognition and behaviour. The grey matter from each activation map was selected using a tissue probability map \cite{vi_segmentation} isolating the tissue of functional origin and processing. The functional distribution of each grey matter voxel was then clustered following the methodology described in \cite{giles_prescriptive_inference} to produce the 16 distinct neural substrates. These 16 distinct loci relate to the functional themes of hearing, language, introspection, cognition, mood, memory, aversion, co-ordination, interoception, sleep, reward, visual recognition, visual perception, spatial reasoning, motor and somatosensory functions. Focal lesions affecting these functional networks can plausibly be simulated to result in deficits within the respective functional theme. Examples of the simulated neural substrates are presented in Figure \ref{fig:neural_susbtrate_fig}. Were these functional divisions secure, we would not need lesion-deficit mapping, but all that is required from the simulated substrates is broad similarity with the spatial structure of real substrates. \newline

\subsection{Simulating functional deficits}
\label{score_simulation}

Having defined the lesions, $\mathbf{X}$, and the substrates, $\mathbf{M}$, we now need to define the deficit function, $\Omega$, that outputs a deficit score $\mathbf{Y}$, given an intersection between substrate and lesion. Deficit functions, just like the neural substrates, are unknown: we therefore need to evaluate a range of possible $\Omega$. \newline

There are many metrics for cognitive and behavioural evaluation \cite{wab_scores, pnt_scores, fast_scores, cat_scores}, with varying scoring systems and values. Given a particular lesion $\mathbf{x}$ and a ground truth neural substrate $\mathbf{m}$, here we calculate the scalar deficit score, $y$, following four different policies.

\subsection{Linear lesion-deficit relation}

This is the simplest modelling assumption, where the cognitive deficit increases linearly with the intersection between the lesion and the neural substrate:

\begin{equation}
    y = \frac{\sum_k x_k m_k}{\sum_k m_k}
\end{equation}

This is the method used for evaluating both the most common multivariate methods in literature \cite{zhang_svr, pustina}.

\subsection{Binary lesion-deficit relation}

A deficit may be scored only as present or absent, especially in the clinical domain\cite{bultmann2014functional}. To account for this scenario we define a binary deficit label as:

\begin{equation}
    y = \begin{cases} 
      1, & \frac{\sum_k x_k m_k}{\sum_k m_k} > T, \\ 
      0, & \text{otherwise}.
   \end{cases}
\end{equation}

That is, the binary outcome is dependent on the proportion of the relevant functional substrate covered by a lesion. We experiment over various ranges of $T$ and evaluate their impact on performance. For all other experiments, we leave $T=0.01$, which implies only 1 \% of the substrate needs to be hit by a lesion for it to generate a deficit.

\subsection{Non-linear lesion-deficit relation}
\label{non_linear_label_sect}

Lesion-deficit relationships are likely to exhibit non-linearities \cite{parashkev_brain}. At the lower end of the function, an increment in lesion volume may cause no or only minimal increase in the magnitude of the deficit owing to redundancy, while at the upper end, saturation is likely once disruption reaches a critical threshold. These considerations motivate the choice of a sigmoid function of the intersection ratio $R = \frac{\sum_k x_k m_k}{\sum_k m_k}$, here the logistic: 

\begin{equation}
    y = \frac{1}{1 + e^{20R - 6}}.
\end{equation}

\begin{figure}[H]
    \centering
    \includegraphics[width=10cm]{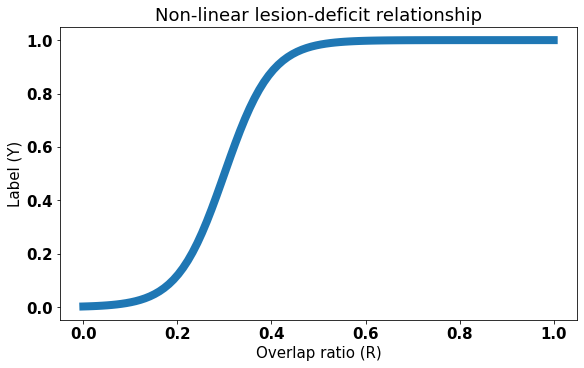}
    \caption[Label]{The chosen non-linear relationship between lesions and functional deficits}
    \label{fig:non_linear_lesion_deficit}
\end{figure}

The plot of the function for the range of possible ratios is presented in Figure \ref{fig:non_linear_lesion_deficit}. \newline

\subsection{Noisy lesion-deficit relation}
\label{noise_explanation}
 
Noise is likely to be present in the labels, from a variety of sources. To account for this, we run experiments with varying quantities of label noise for all the previously described score simulation methods. The types of noise used are as follows:
\begin{itemize}
    \item \textbf{Binary noise}: For each label, $y$, we flip its value with a probability $p$. This type of noise accounts for variations in the substrate across the population, and the expression and recording of the associated deficit. We examine values in the range $p \in [0, 0.3]$.
    \item \textbf{Linear noise}: For each label, $y$, we take noise samples from the uniform distribution $\epsilon \sim \mathcal{U}(0, 1)$ and use the convex combination $(1-\alpha)y + \alpha \epsilon$, where $0 \leq \alpha \leq 1$, as the final label. This is the setup used in \cite{pustina}, which the authors found to result in simulated cognitive scores most similar to real data. We examine a range of possible noise levels with $\alpha \in [0, 0.5]$
\end{itemize}

\subsection{Heterogeneous lesion-deficit relation}
\label{heterogeneity_explanation}
Patient populations may be heterogeneous, systematically varying in the organisation of the underlying substrate\cite{parashkev_brain}. Such heterogeneity may be manifest in the substrate alone, and so impossible to filter \textit{a priori}. It differs from noise in being potentially learnable with a model of sufficient expressivity. Here we simulate heterogeneity by randomly drawing the ground truth for a lesion-deficit relation from one of two different substrates, evenly split across the cohort. The correct inference at the population level should include \textit{both} substrates, even if each is specific to its own subpopulation.     

\subsection{Validation and Calibration sets}

Having defined our lesion dataset, $\mathbf{x}$, ground truth substrates, $\mathbf{m}$, and deficit functions through which to obtain $\mathbf{y}$ we can now define the training, validation and test splits. \newline

For the task of evaluating substrate recovery, the test set is simply the true functional substrate map. Therefore, none of the methods require allocating data to a test set. Note that none of the methods are exposed to this map at any time during training or hyperparameter optimization, for it is definitionally not available in the real setting.  All multivariate models, ours included, require a \textit{predictive} validation set, composed only of lesion maps and their associated deficits, either to avoid overfitting on the training data (SVR-LSM and DLM), or to choose optimal hyperparameters (SCCAN and SVR-LSM). We set aside 10\% of the lesion data for the validation set and we ensure that no patients are shared across training and validation sets. \newline

Furthermore, for our model, we split the 10\% into two halves, one for the validation set, which we use to select the best model and another half for the calibration set, which is used to select the threshold to use for the binarisation of the recovered map $\hat{\mathbf{m}}$. VLSM does not require a validation set and therefore uses all the data provided for inference. \newline

\subsection{Binary lesion deficit experiments}
\label{intro_to_binary}

 To maximise statistical efficiency, lesion-deficit studies seek to achieve 1:1 ratios of positive to negative labels \cite{rct_ratios}. Most studies operate within low data regimes, employing of the order of 100 labels \cite{svrlsm_eval, bates_vlsm, pustina}. Our experiments use stratified sampling from the 5500 available lesions and simulated labels to ensure a 1:1 label ratio where the cardinality of the data is less than 500. \newline

Owing to the distinctive anatomical distribution of acute ischaemic lesions, we cannot assure a ratio of 1:1 for experiments that use more than 500 labels, because negative labels are more numerous than positive ones. For such cases, we conduct stratified sampling from all our data, first attempting a 40/60 positive/negative ratio, followed by 30/70, 20/80 and 10/90 ratios. If we cannot ensure at least a 10/90 ratio the experiment is excluded from analysis. \newline

\subsection{Fixed binary threshold}
\label{fixed_threshold_binary}

\begin{figure}[H]
\hspace*{-1cm}
    \centering
    \includegraphics[width=12cm]{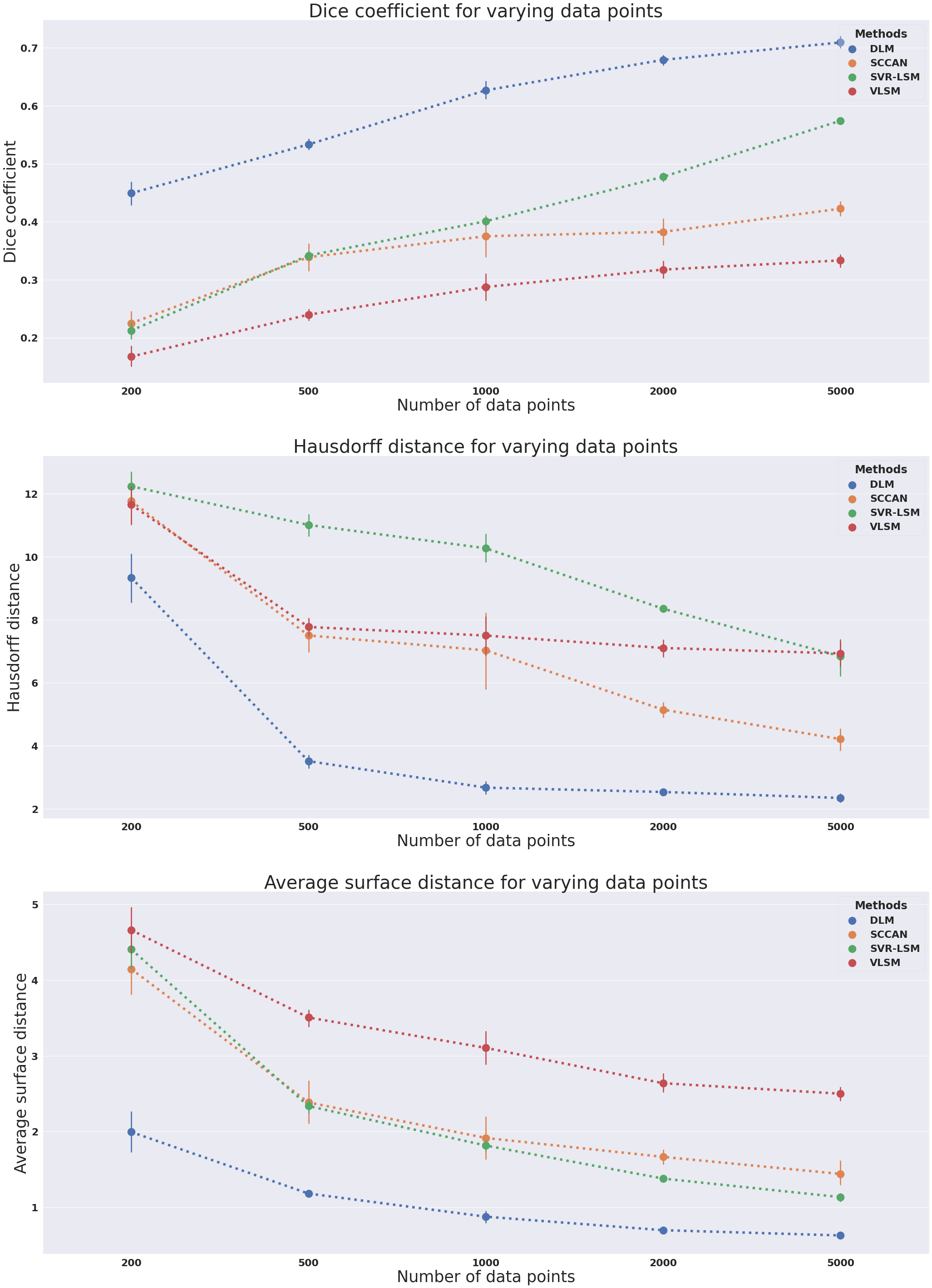}
    \caption[Model performances on the task of substrate recovery given binary lesion deficit]{Model performance on the task of substrate recovery given binary lesion deficit labels for varying amounts of data. The error bars correspond to 1 standard deviation. Dice (top), higher is better. Hausdorff distance (middle), lower is better. Average surface distance (bottom), lower is better.}
    \label{fig:varying_n_binary}
\end{figure}

\begin{figure}[H]
\hspace*{-0.5cm}
    \centering
    \includegraphics[width=13cm]{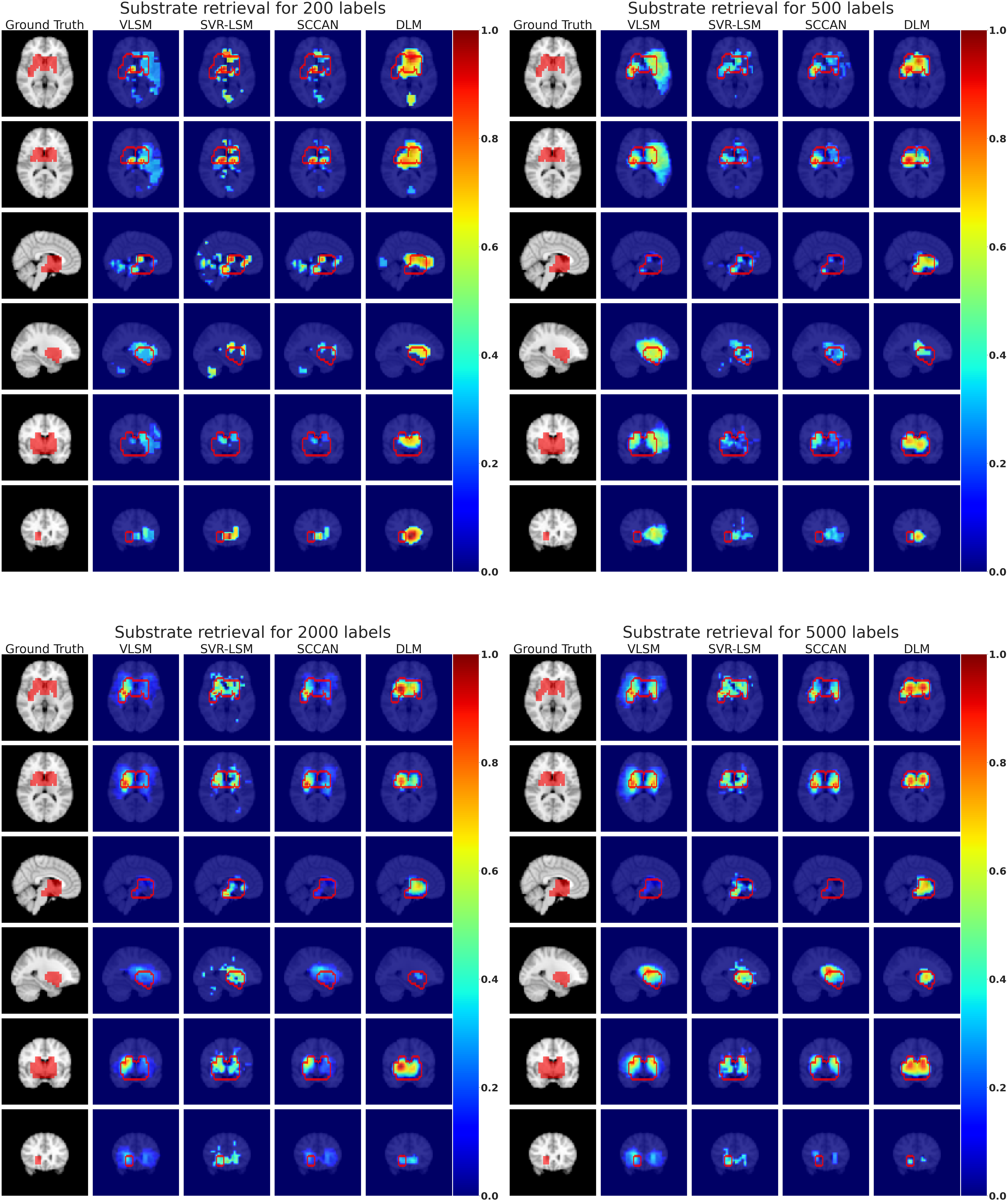}
    \caption[Example substrate recoveries for binary labels 1]{Example substrate recoveries for binary labels and varying amounts of available data for VLSM, SVR-LSM, SCCAN and DLM.}
    \label{fig:substrate_recover_varying_binary_n_11}
\end{figure}

\begin{figure}[H]
\hspace*{-0.5cm}
    \centering
    \includegraphics[width=13cm]{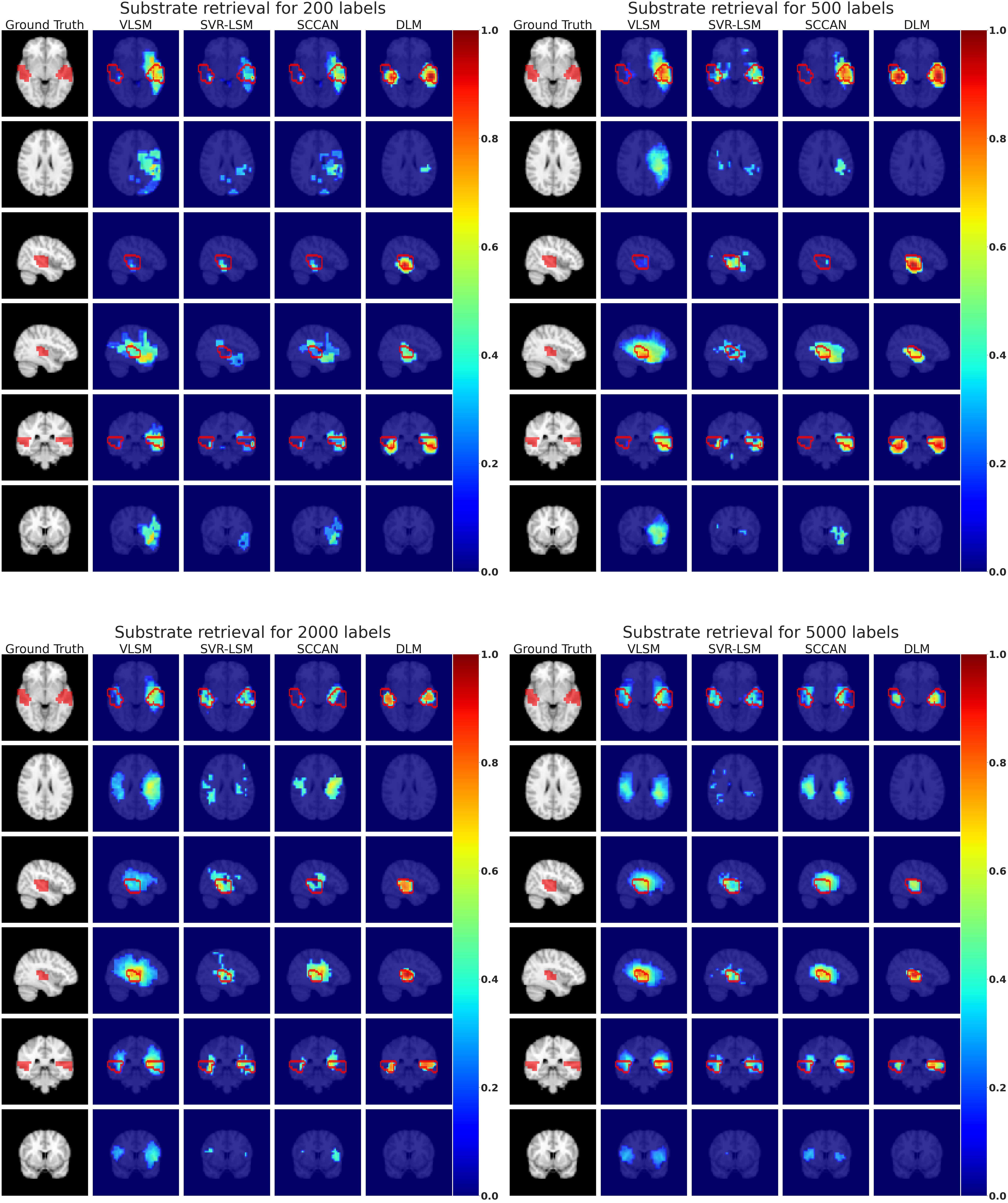}
    \caption[Example substrate recoveries for binary labels 2]{Example substrate recoveries for binary labels and varying amounts of available data for VLSM, SVR-LSM, SCCAN and DLM.}
    \label{fig:substrate_recover_varying_binary_n_1}
\end{figure}

Using a threshold of 1\%, i.e. $T=0.01$, the results obtained for each method on all metrics are displayed in Figure \ref{fig:varying_n_binary} and example substrate recoveries are presented in Figures \ref{fig:substrate_recover_varying_binary_n_11} and \ref{fig:substrate_recover_varying_binary_n_1}. This is the first time in literature that the both SVR-LSM and SCCAN have been benchmarked on the task of inference with binary labels. \newline

In terms of Dice, all multivariate methods consistently outperform VLSM across all data regimes (Figure \ref{fig:varying_n_binary}). SVR-LSM outperforms SCCAN at $N \geq 1000$, which implies it has great success in reducing both false positive and false negatives, however, it always underperforms in terms of HD, implying that the false positives it predicts are further way from the ground truth. SCCAN consistently outperforms VLSM on all metrics, and could therefore be the better of the traditional multivariate methods for binary tasks. \newline

DLM, outperforms all other methods, across all metrics and all data regimes. The superiority is substantial even under low data regimes. The slope on the metric lines for both HD and ASD is highest at $200\leq N\leq500$, whereas it remains relatively constant for Dice throughout. Coupled with the qualitative analysis of Figures \ref{fig:substrate_recover_varying_binary_n_11} and \ref{fig:substrate_recover_varying_binary_n_1} this implies the model is better at removing false positives even with $N\leq500$. By using an unseen data distribution for calibration, as opposed to performing inference entirely on the same data distribution as with VLSM, SCCAN and SVR-LSM, false positive voxels are more likely to be avoided. This is especially relevant under lower data regimes, where spurious patterns are more likely to be learnt, hence the flattening of the slope for both HD and ASD as $N$ increases. \newline

The linear improvement in Dice is arguably due to DLM learning the lesion distribution $P(\mathbf{X})$. As has been noted in \cite{yee_parashkev}, stereotyped patterns of damage driven by the structure of the pathological process lead to mislocalisation where correlations across loci are inadequately modelled. Hence, DLM by learning the joint distribution of lesions and associated deficits $P(\mathbf{X}, \mathbf{Y})$, succeeds in reducing the stereotyped pattern bias. Note that despite being the best performing model, due to the topological complexity of our substrates and lesions, DLM yields Dice, HD and ASD scores short of those achievable in semantic segmentation \cite{unetr}, where training on ground truths is possible. This further underscores the need for larger data regimes in the field of lesion deficit mapping. \newline

\subsection{Ablation study}

A theoretical justification for the choice of modelling framework was given in \S \ref{vae_ldm_theory}. In this section we provide an empirical justification for our chosen framework through ablation studies of our selected neural architecture. The studied ablations are as follows:
\begin{itemize}
    \item U-Net framework: We use a U-Net \cite{unet} network with an equivalent number of parameters to our model that optimises the loss $\log P(\mathbf{Y}=\mathbf{y}|\mathbf{X}=\mathbf{x}, \mathbf{M}=\mathbf{m})$ directly. In other words, the model's training objective is to output a continuous deterministic map $\mathbf{m}$ that maximises the predictive accuracy on the training and validation set. This map $\mathbf{m}$ is then thresholded with the calibration set and evaluated against the ground truth neural substrate.
    \item Autoencoder (AE) framework: Rather than using the VAE framework, we use a deterministic autoencoder \cite{andrew_ng_autoencoder} to model the same loss our model optimises except without the KL divergence. 
    \item VAE of label distribution only: Rather than modelling the joint distribution, $P(\mathbf{X}, \mathbf{Y})$, we now model only the distribution of labels $P(\mathbf{Y})$. We optimise $ \mathbb{E}_{\mathbf{z}\sim Q_{\boldsymbol{\phi}}} [\log P_{\boldsymbol{\gamma}}(\mathbf{Y}|\mathbf{X},\mathbf{Z})] - D_{KL}(Q_{\boldsymbol{\phi}}(\mathbf{Z}|\mathbf{X}, \mathbf{Y}) || P_{\boldsymbol{\psi}}(\mathbf{Z}))$. 
\end{itemize}

We evaluate all listed ablations for $N=5000$ and all possible ground truth neural substrates. We also include the best performing baseline, SVR-LSM, as the nearest reference point. The results are presented on Table \ref{ablations_table}.

\begin{table}[h]
\centering
\begin{tabular}{llll} \toprule
 Model & Dice & Hausdorff & ASD \\ \midrule
 U-Net & 0.475 $\pm$ 0.029 & 4.59 $\pm$ 0.13 & 1.54 $\pm$ 0.05 \\
 AE &  0.481 $\pm$ 0.032 & 4.55 $\pm$ 0.12 & 1.53 $\pm$ 0.06 \\
 SVR-LSM &  0.578 $\pm$ 0.008 & 4.09 $\pm$ 0.15 & 1.23 $\pm$ 0.05 \\
 DLM-labels-only &  0.628 $\pm$ 0.016 & 3.21 $\pm$ 0.09 & 1.06 $\pm$ 0.08 \\
 \textbf{DLM} & \textbf{0.715} $\pm$ 0.014 & \textbf{2.18} $\pm$ 0.08 & \textbf{0.52} $\pm$ 0.04 \\
 \bottomrule
\end{tabular}
\caption{Comparative impact of model ablation on inferential performance for binary labels and $N=5000$. Best performance is highlighted in bold.}\label{ablations_table}
\end{table}

Both deterministic methods, U-Net and AE, perform the worst, illustrating the importance of learning a distribution over possible neural substrates rather than a single deterministic map. Note that both ablated methods underperform SVR-LSM, despite the scale of training data, and the superiority of these methods on conventional discriminative tasks \cite{unet}. This highlights the importance of carefully considering the specific problem formulation when extending neural networks to the task of inference (as opposed to prediction). When training DLM only on the label distribution, (DLM-labels-only), there is a substantial deterioration in performance. This finding reflects the inevitable influence of both lesion and functional structure in determining the inferrability of lesion-deficit relations \cite{tianbo_ldm, mah_multivariate, yee_parashkev}. \newline

Note that ablation studies are substantially less important here than in conventional predictive tasks, for in the absence of a ground truth the adequacy of any minimal architecture cannot be guaranteed. The primary guide to architectural development must be the model properties outlined in \S \ref{intro}, whose desirability is given \textit{a priori}. It would be hazardous to \textit{fit} an architecture to simulated ground truths even if we may use reasonably \textit{evaluate} architectures in this---the only empirically accessible---way. For this, and reasons of computational cost, we do conduct ablated replications of every single experimental scenario.     

\subsection{Learnt lesion distribution}

As explained in \S \ref{vae_ldm_theory}, DLM is concerned with modelling the joint distribution $P(\mathbf{X}, \mathbf{Y})$. The foregoing experiments focus on inference to the simulated neural substrate, but a critical component of the model's inferential ability is its ability to capture the lesion distribution $\log P_{\boldsymbol{\theta}}(\mathbf{X}|\mathbf{Z})$. Hence, in this section we focus on verifying that this distribution is indeed faithfully modelled. \newline

We use models trained on binary lesion deficits, similarly to the previous section to do this verification. We find that although modelling the lesion likelihood $\log P_{\boldsymbol{\theta}}(\mathbf{X}|\mathbf{Z})$ helps with achieving more accurate $\log P_{\boldsymbol{\gamma}}(\mathbf{Y}|\mathbf{X},\mathbf{Z})$, the nature of $\mathbf{Y}$ has very little effect on modelling $\log P_{\boldsymbol{\theta}}(\mathbf{X}|\mathbf{Z})$ (as expected). Hence, the conclusions achieved in this section hold for all subsequent experiments, regardless of nature of the lesion-deficit relationship. \newline

\begin{figure}[H]
\hspace*{-0.5cm}
    \centering
    \includegraphics[width=15cm]{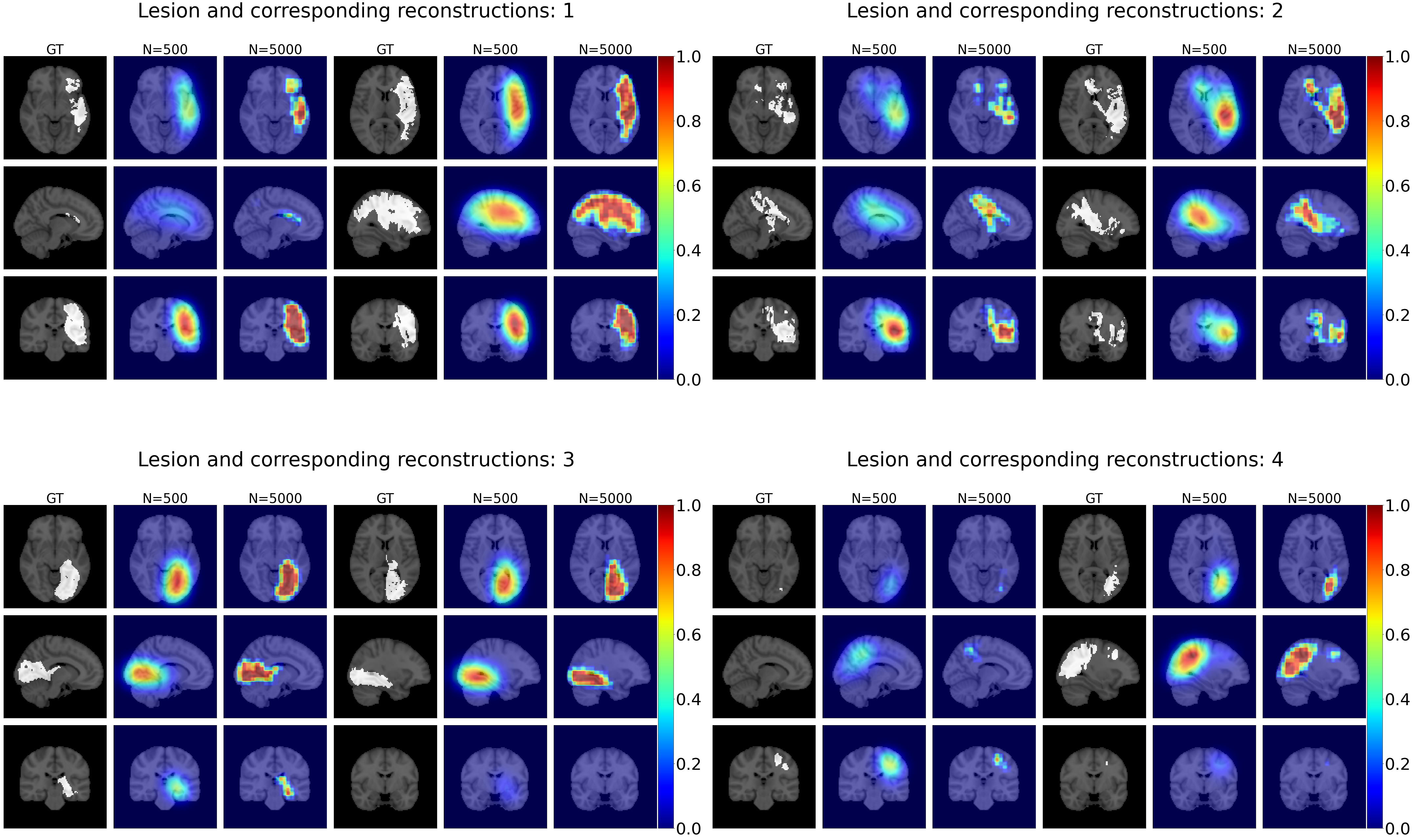}
    \caption[Example DLM lesion reconstructions at various data amounts]{Lesion examples and their corresponding DLM reconstructions at $N=500$ and $N=5000$.}
    \label{fig:lesion_reconstructions}
\end{figure}

\begin{figure}[H]
\hspace*{-1cm}
    \centering
    \includegraphics[width=15cm]{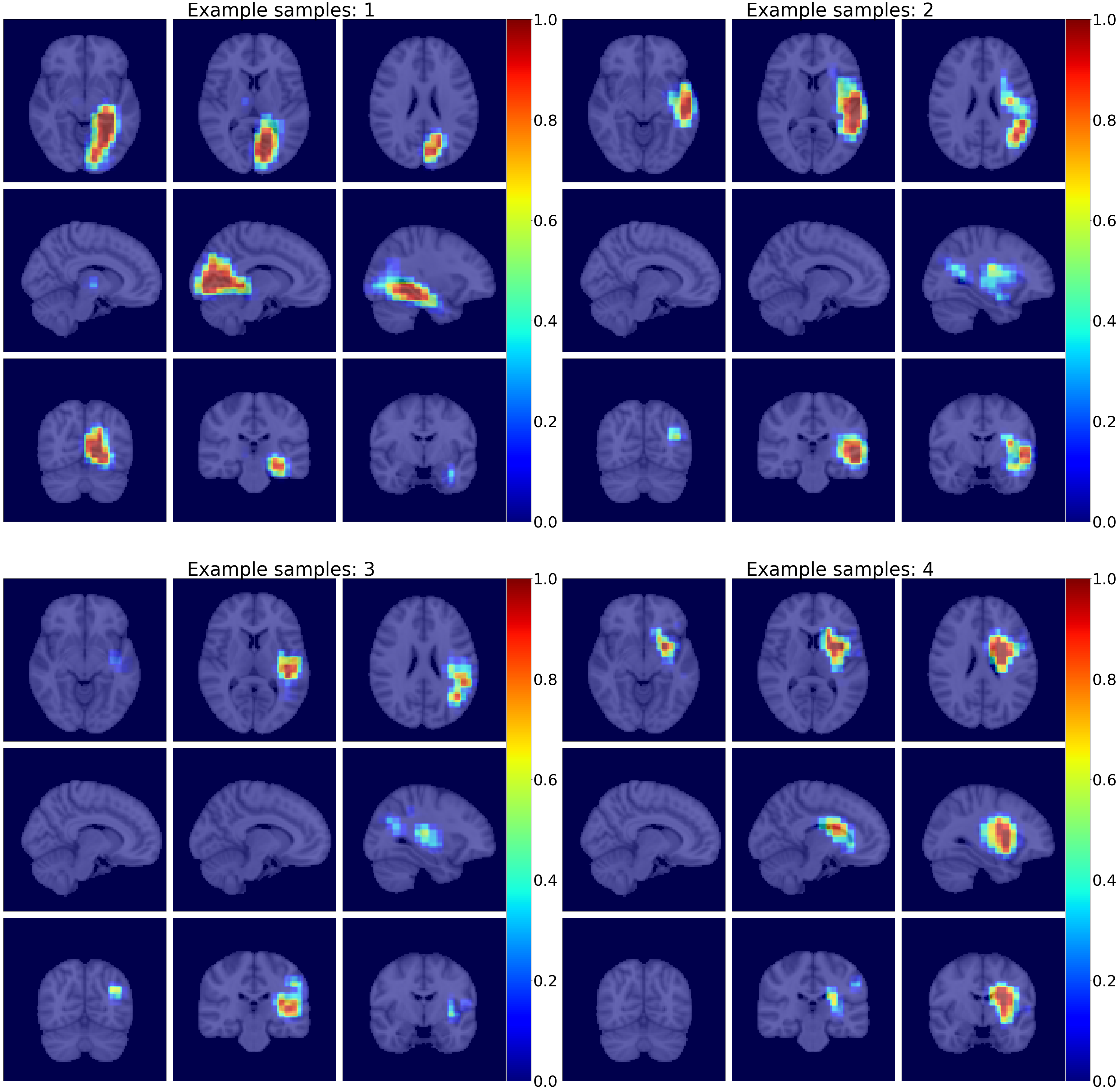}
    \caption[Example DLM samples lesions]{Example sampled lesions from the DLM model}
    \label{fig:vae_lesion_samples}
\end{figure}

In order to evaluate the reconstruction quality, for these sets of experiments alone we use a held out test set of 100 lesions. In Figure \ref{fig:lesion_reconstructions} we present some example lesions from this held out test set and the respective DLM reconstructions at both $N=500$ and $N=5000$. In Table \ref{recons_table} we present the metrics obtained by the DLM reconstructions on the test set lesions. We note that despite being a slightly different problem, we obtain metrics that are comparable to state-of-the-art brain lesion segmentation \cite{unetr}, when using larger numbers of data points $N \geq 2000$. Dice and ASD go down smoothly with increases in available data; however, the Hausdorff distance decreases significantly less, since it is the metric most sensitive to details. Its reduction being less attenuated is a healthy indicator of the model's ability to avoid overfitting in order to reconstruct lesions accurately, learning the general topology and distribution of ischaemic stroke lesions instead. Finally, in Figure \ref{fig:vae_lesion_samples} we present some sample lesions obtained by sampling $\mathbf{z} \sim N(0,1)$ and then decoding with $\boldsymbol{\theta}(\mathbf{z})$. The sampled lesions exhibit irregular contours, but they conform qualitatively to the expected topology of ischaemic stroke lesions described in \cite{stroke_lesion_anatomy}. \newline

\begin{table}[h]
\centering
\begin{tabular}{llll} \toprule
 Data Points (N) & Dice & Hausdorff & ASD \\ \midrule
 N=200 &  0.302 $\pm$ 0.05 & 8.54 $\pm$ 3.24 & 4.42 $\pm$  0.36 \\
 N=500 &  0.388 $\pm$ 0.05 & 8.44 $\pm$ 3.13 & 3.22 $\pm$  0.28 \\
 N=1000 &  0.411 $\pm$ 0.06 & 8.30 $\pm$ 3.03 & 2.93 $\pm$  0.27 \\
 N=2000 &  0.512 $\pm$ 0.04 & 8.02 $\pm$ 2.98 & 1.99 $\pm$  0.19 \\
 N=5000 &  0.713 $\pm$ 0.03 & 7.78 $\pm$ 2.82 & 0.672 $\pm$  0.03 \\
 \bottomrule
\end{tabular}
\caption{Evaluation of reconstruction of test set lesions}\label{recons_table}
\end{table}

\subsection{Varying binary thresholds}

\begin{figure}[H]
\hspace*{-1cm}
    \centering
    \includegraphics[width=14cm]{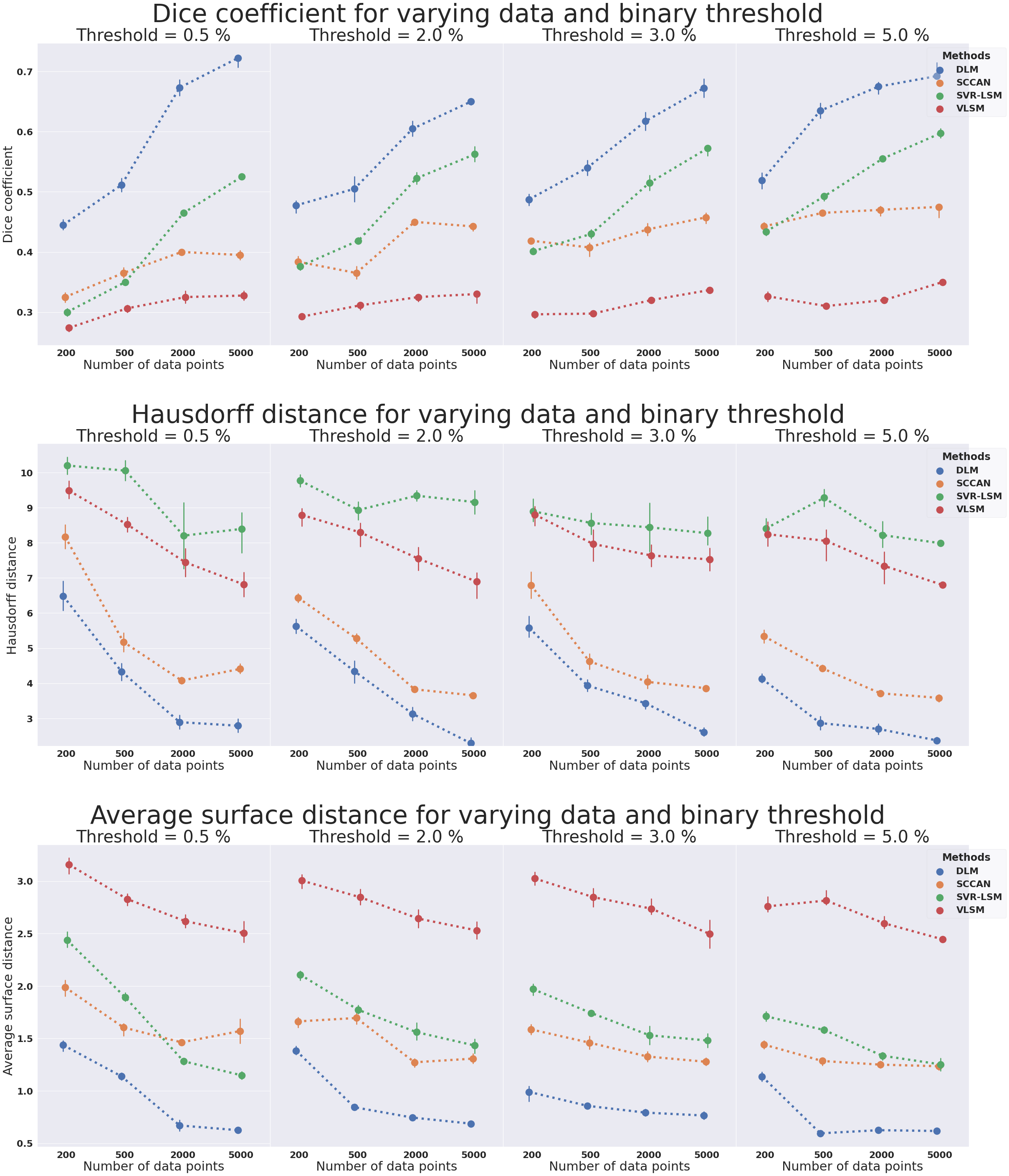}
    \caption[Model performance on the task of substrate recovery given binary lesion deficit for varying thresholds]{Model performance on the task of substrate recovery given binary lesion deficit labels for varying amounts of data and varying binary thresholds. The error bars correspond to 1 standard deviation. From left to right we vary the binary threshold, from top to bottom we vary the metric to be gauge. Dice (top), higher is better. Hausdorff distance (middle), lower is better. Average surface distance (bottom), lower is better.}
    \label{fig:varying_n_binary_threshold_1}
\end{figure}

\begin{figure}[H]
\hspace*{-0.5cm}
    \centering
    \includegraphics[width=15cm]{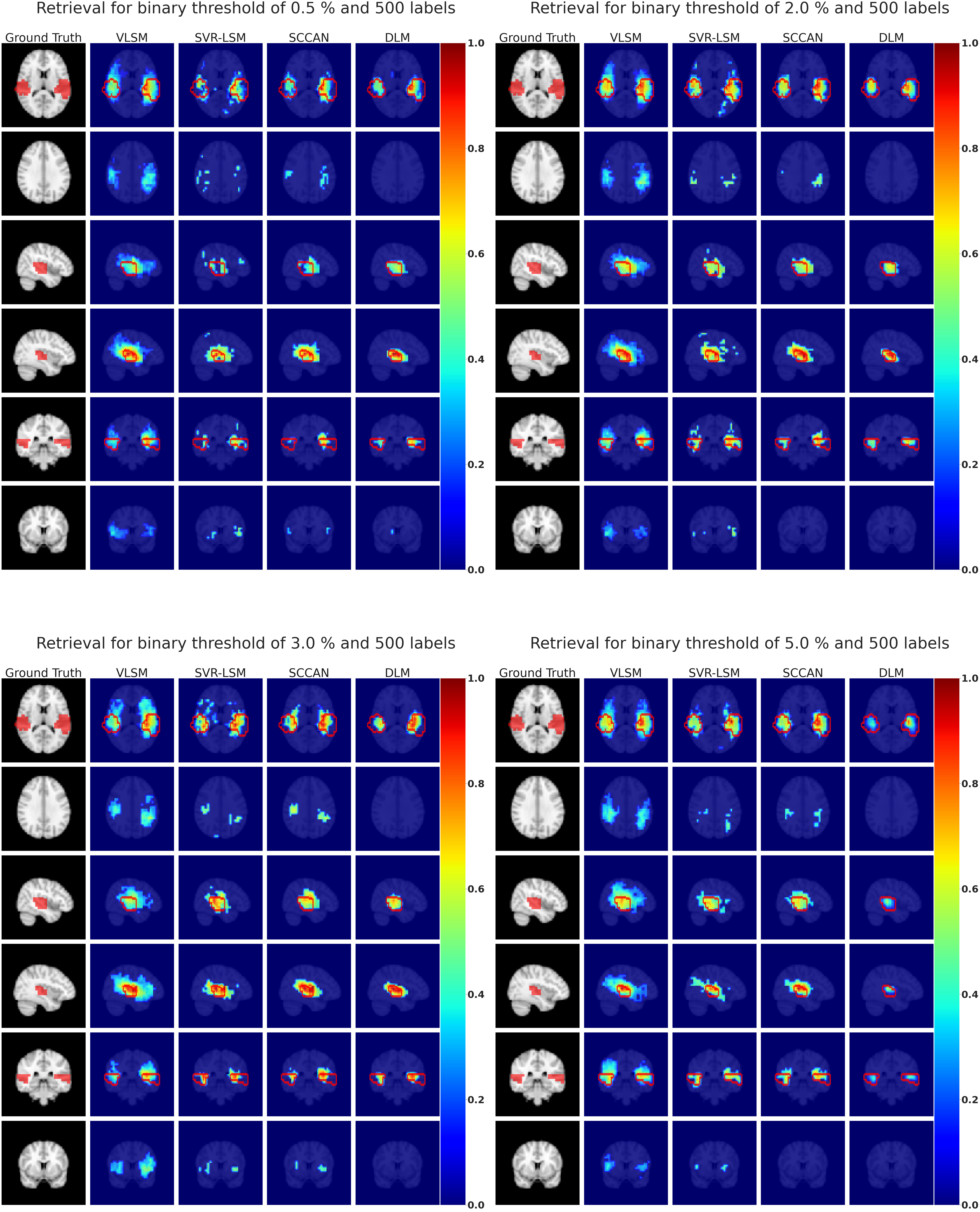}
    \caption[Example substrate recoveries for binary deficit labels with varying thresholds]{Example substrate recoveries for binary deficit labels with varying thresholds and 500 labels for VLSM, SVR-LSM, SCCAN and DLM.}
    \label{fig:substrate_recover_varying_threshold_n_1}
\end{figure}

In the experiments presented in \S \ref{fixed_threshold_binary} we used a fixed threshold $T=0.01$; in this section we explore how model performance changes as we change the threshold. We choose a threshold range $T \in [0.005, 0.05]$ that allows for ratios of at least 10/90 (negative/positive) to be maintained at $N=5000$ for at least 4 of the neural substrates described in \S \ref{functional_neural substrate_section}. At any higher threshold, too few positive labels are available, reducing the realism of the simulations. Model performances for varying thresholds and amounts of data are shown in Figure \ref{fig:varying_n_binary_threshold_1} and example substrate recoveries for various thresholds at $N=500$ are presented in Figure \ref{fig:substrate_recover_varying_threshold_n_1}. \newline

The absence of substantial changes in DLM model performances with varying thresholds indicates stability under shifting class imbalance, and leaves the conclusions of \S \ref{fixed_threshold_binary} remain unchanged. SCCAN and SVR-LSM show greater sensitivity to threshold, especially at lower sample numbers. However, the final performance at $N\geq2000$ remains relatively unchanged, since neither method can leverage negatively labelled lesions effectively. This phenomenon is qualitatively illustrated in Figure \ref{fig:substrate_recover_varying_threshold_n_1} where as the threshold is raised, we see a simultaneous increase of false positives and decrease of false negatives for all models except for DLM.\newline

\subsection{Varying amounts of symmetric noise}
\label{experiments_bin_noise}

\begin{figure}[H]
\hspace*{-1cm}
    \centering
    \includegraphics[width=14cm]{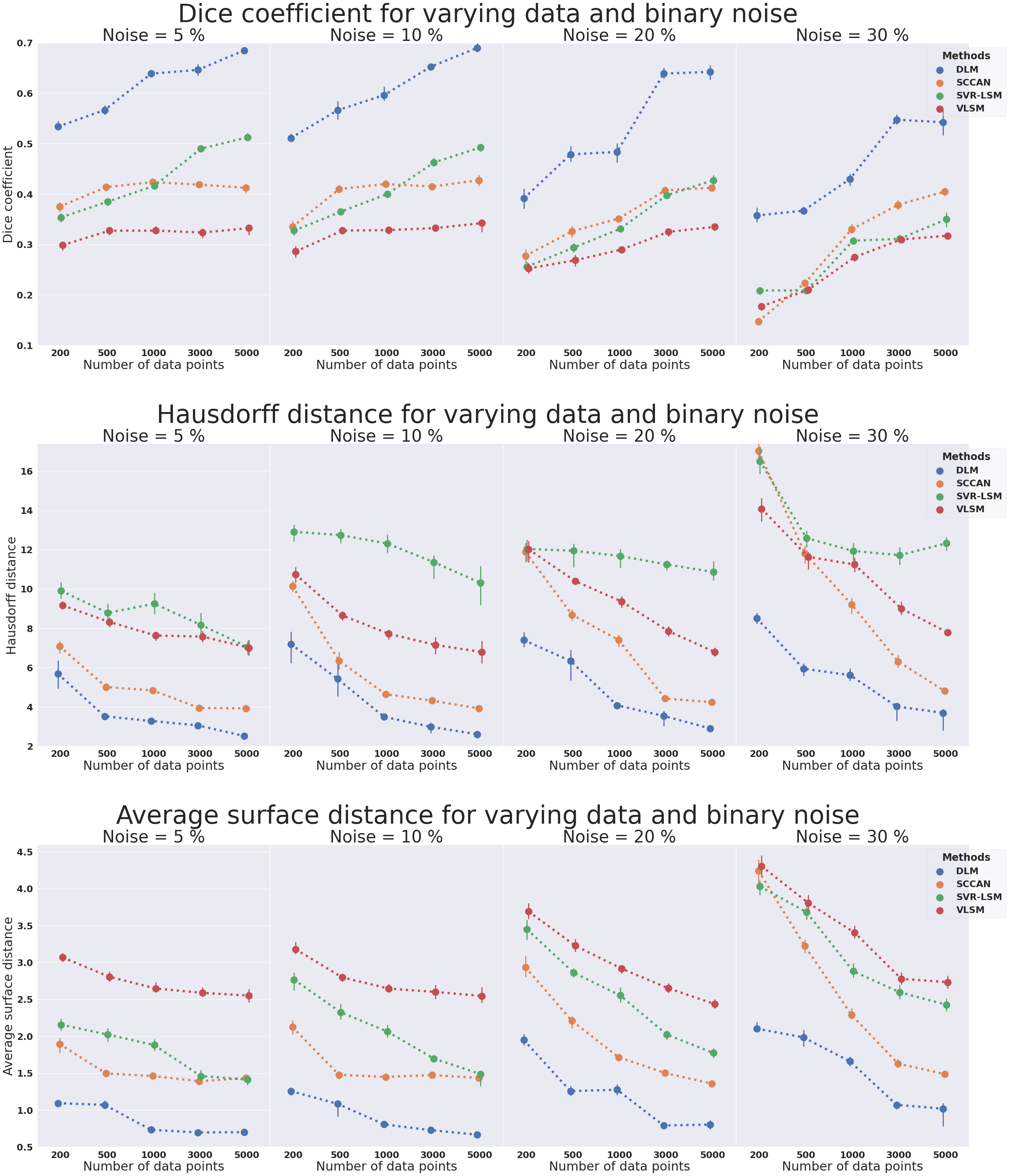}
    \caption[Model performance on the task of substrate recovery given binary lesion deficit labels and varying amounts of binary noise]{Model performance on the task of substrate recovery given binary lesion deficit labels for varying amounts of data and varying amounts of binary noise. The error bars correspond to 1 standard deviation. From left to right we vary the amount of noise, from top to bottom we vary the metric to be gauge. Dice (top), higher is better. Hausdorff distance (middle), lower is better. Average surface distance (bottom), lower is better.}
    \label{fig:varying_n_binary_noise_1}
\end{figure}

\begin{figure}[H]
\hspace*{-1cm}
    \centering
    \includegraphics[width=15cm]{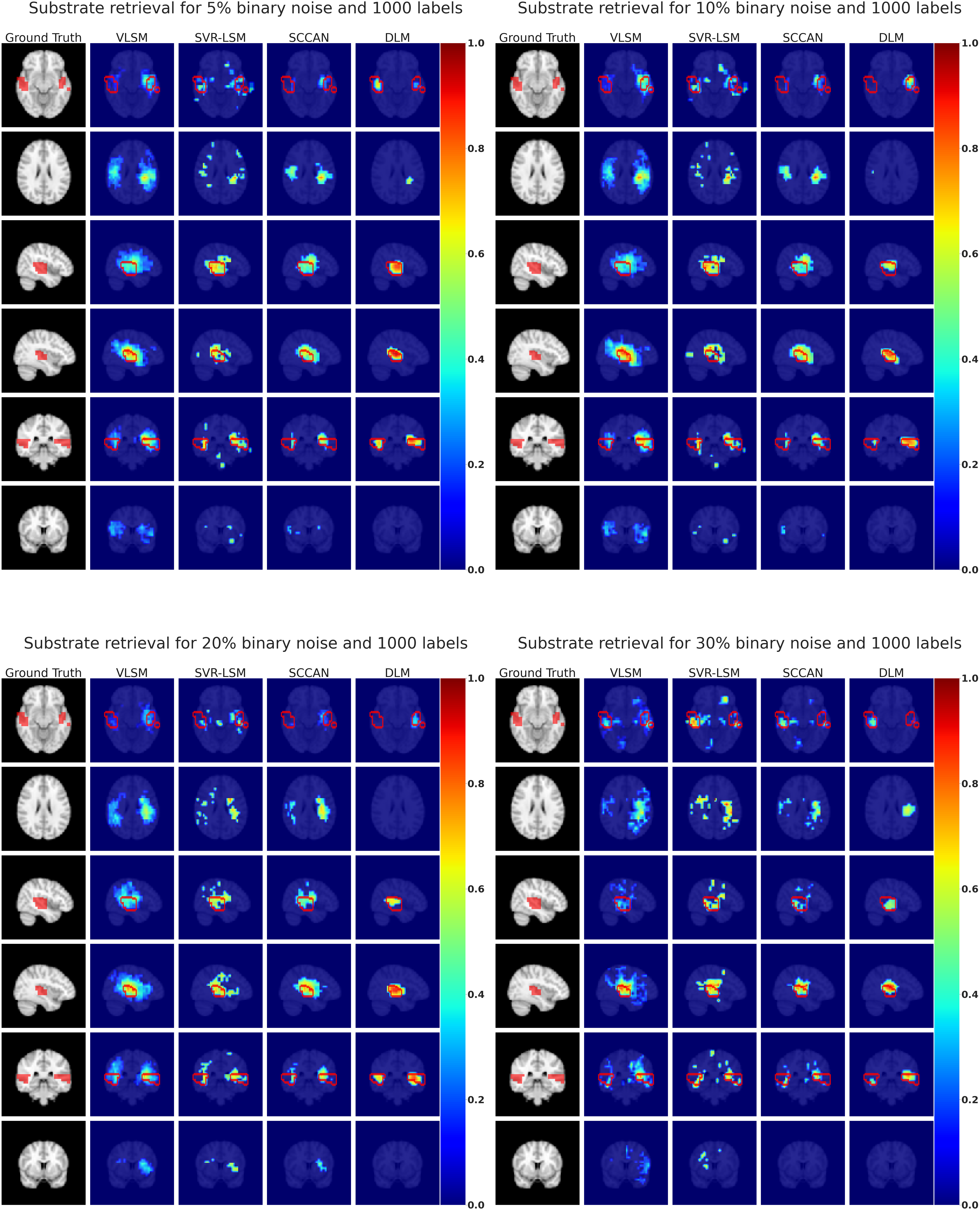}
    \caption[Example substrate recoveries for for binary deficit labels with varying amounts of binary symmetric noise 1]{Example substrate recoveries for for binary deficit labels with varying amounts of binary symmetric noise and 1000 labels for VLSM, SVR-LSM, SCCAN and DLM.}
    \label{fig:substrate_recover_varying_binary_noise_1}
\end{figure}

\begin{figure}[H]
\hspace*{-1cm}
    \centering
    \includegraphics[width=15cm]{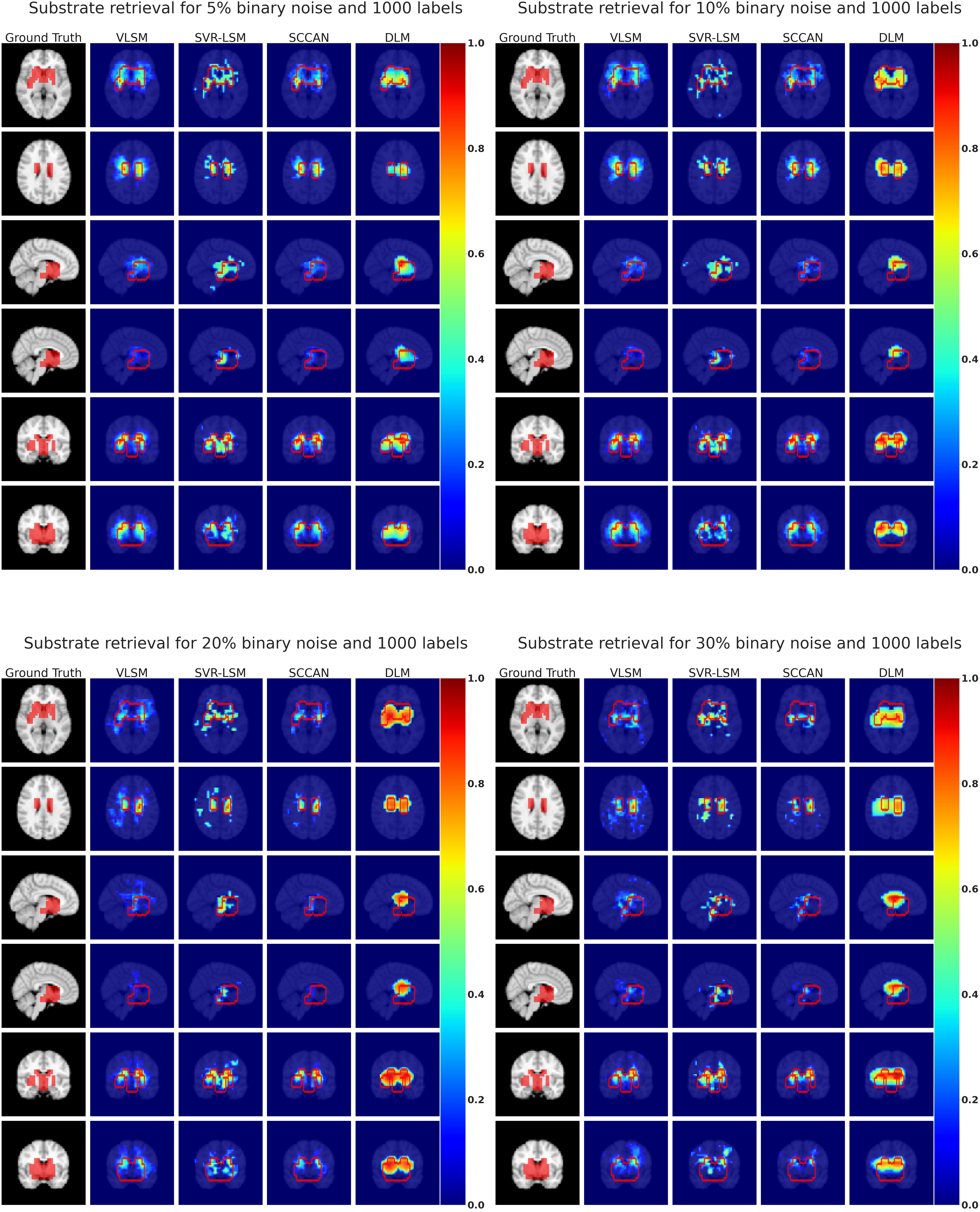}
    \caption[Example substrate recoveries for for binary deficit labels with varying amounts of binary symmetric noise 2]{Example substrate recoveries for for binary deficit labels with varying amounts of binary symmetric noise and 1000 labels for VLSM, SVR-LSM, SCCAN and DLM.}
    \label{fig:substrate_recover_varying_binary_noise_11}
\end{figure}

In the previous set of experiments, we used clean labels which is an infrequent scenario in the real world. As described in \S \ref{noise_explanation} we now randomly flip the binary labels with varying probability $p$. In this section we explore how model performance changes as we increase $p$. The model performances for varying amounts of noise and amounts of data are shown in Figure \ref{fig:varying_n_binary_noise_1} and example substrate recoveries for various amounts of noise at $N=1000$ are presented in Figures \ref{fig:substrate_recover_varying_binary_noise_1} and \ref{fig:substrate_recover_varying_binary_noise_11}. \newline

As expected, there is a deterioration in performance across all metrics as noise is increased. Qualitatively one can observe in Figures \ref{fig:substrate_recover_varying_binary_noise_1} and \ref{fig:substrate_recover_varying_binary_noise_11} that this deterioration seems to stem equally from increases in false positives as well as increases in false negatives. As in the previous sets of experiments, SVR-LSM still outperforms SCCAN with regards to Dice, but however, as the noise increases past the 10\% mark, it is noticeable that in general SCCAN starts to outperform SVR-LSM on all metrics. Out of the baselines we compare DLM to, VLSM seems to be the most resilient to noise, which is expected, since it also the model with highest bias. \newline

DLM consistently outperforms all baselines on all metrics, overfitting less to the training and validation data than other models despite having more parameters. Even at the lowest data scales $N \leq 500$, DLM outperforms all other models. All multivariate models that were tested use some form of  regularisation (either L1 or L2) and therefore that cannot be an explanation for DLM's superior noise resistance. \newline

Following \cite{variational_robust}, we suggest that our model's resilience to noise is explained by its use of variational modelling: the latent variables $\mathbf{Z}$ used to model the likelihood $\log P_{\boldsymbol{\gamma}}(\mathbf{Y}|\mathbf{X},\mathbf{Z})$ are sampled from a learnt multi-dimensional mean, $\boldsymbol{\mu}$ and standard deviation, $\boldsymbol{\sigma}$. This allows the label noise to be accounted for via $\boldsymbol{\sigma}$. Furthermore, stochastically sampling the latent variables from a normal distribution acts as a form of data augmentation, promoting resistance to label noise and overfitting \cite{latent_space_augmentation}. Lesion-deficit data is conventionally impossible to augment, since augmentation changes the semantics of the image; however, latent space augmentation allows one to circumvent these limitations, allowing DLM to benefit from the robustness and performance enhancement data augmentations bring.\newline

\subsection{Continuous lesion deficits}

\subsection{Noise-free linear relationship}

\begin{figure}[H]
    \centering
    \includegraphics[width=12cm]{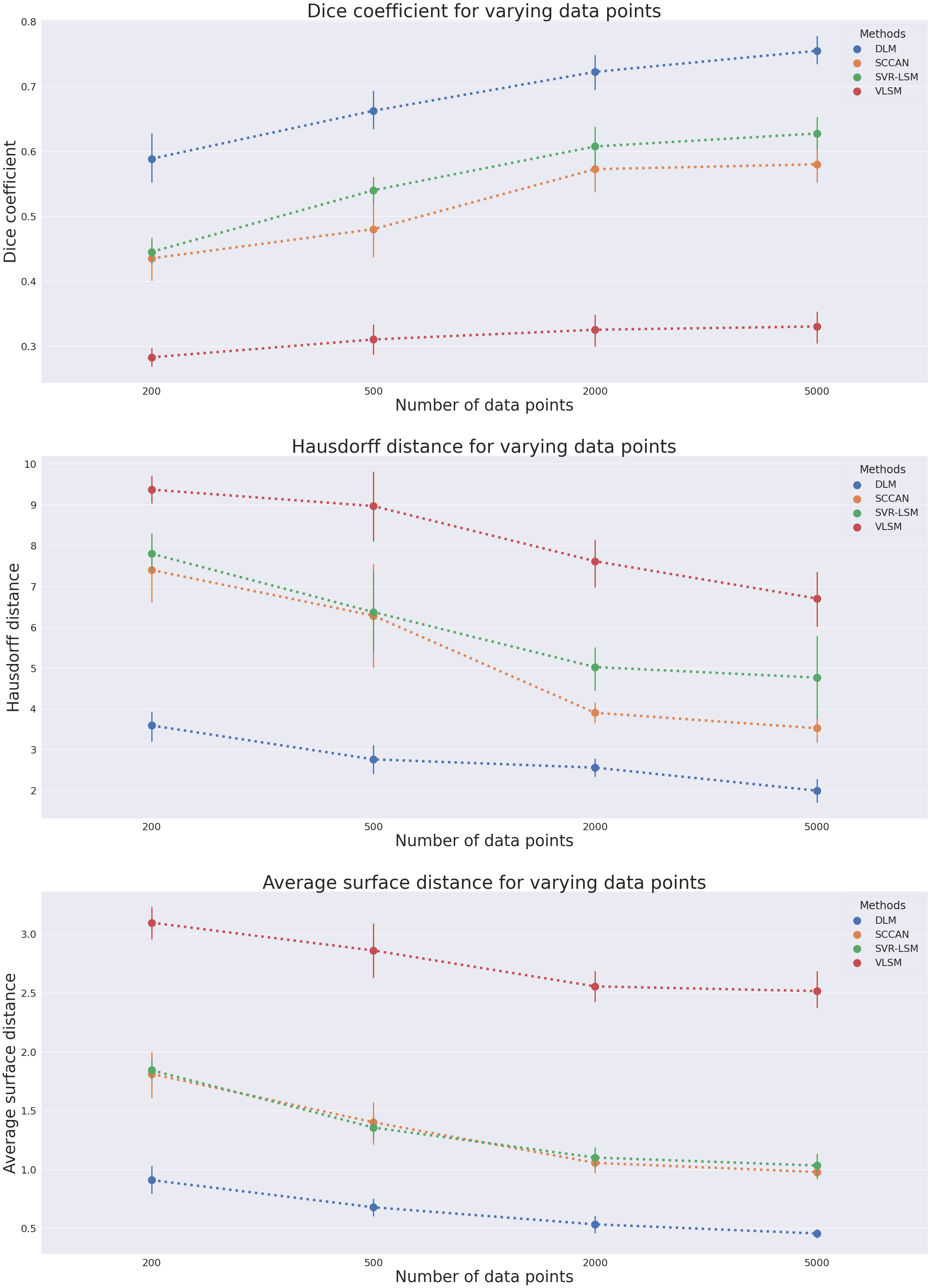}
    \caption[Model performance on the task of substrate recovery given continuous lesion deficit labels]{Model performance on the task of substrate recovery given continuous lesion deficit labels for varying amounts of data. The error bars correspond to 1 standard deviation. Dice (top), higher is better. Hausdorff distance (middle), lower is better. Average surface distance (bottom), lower is better.}
    \label{fig:varying_n_continuous}
\end{figure}

\begin{figure}[H]
\hspace*{-1cm}
    \centering
    \includegraphics[width=15cm]{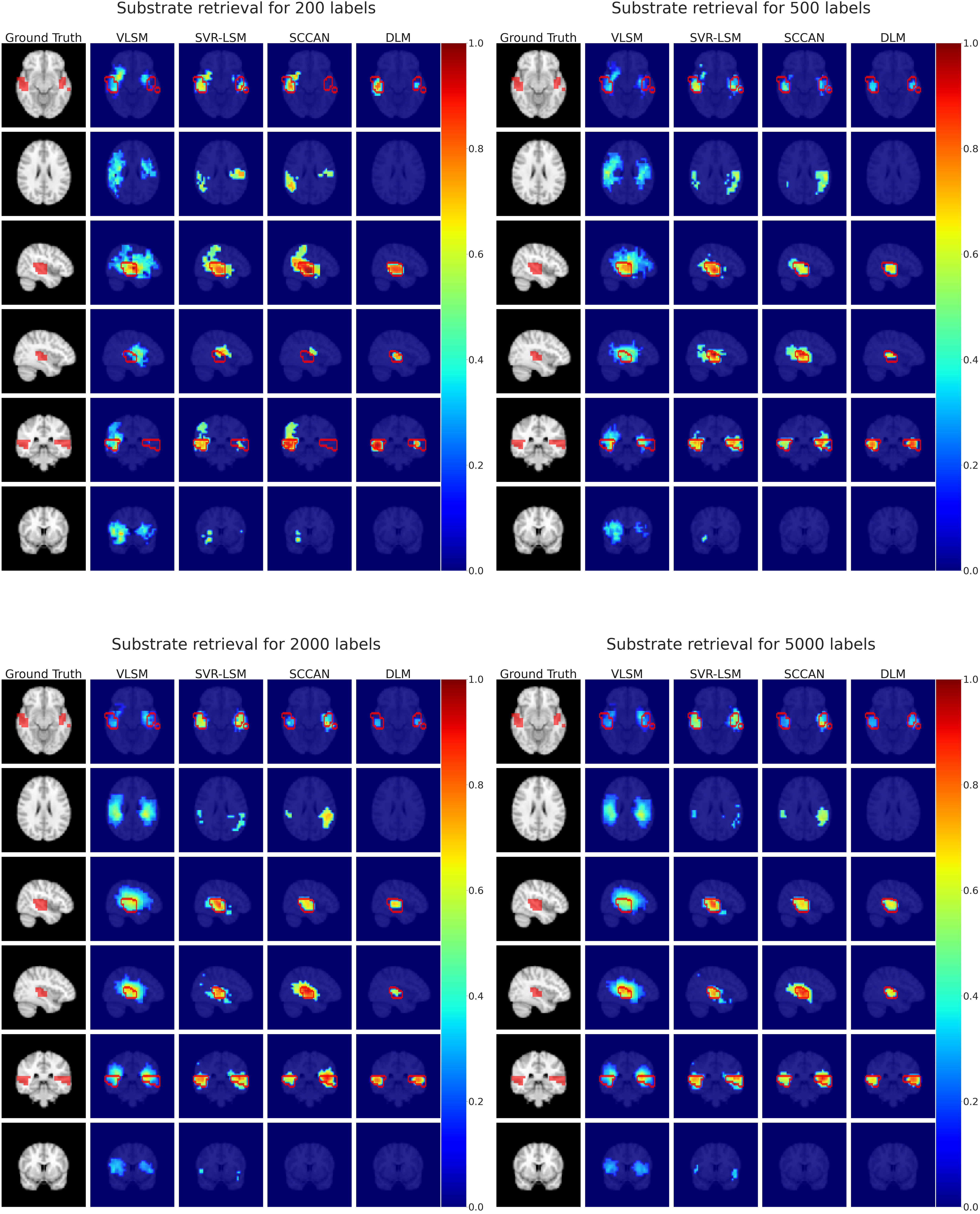}
    \caption[Example substrate recoveries for continuous deficit labels]{Example substrate recoveries for continuous deficit labels and varying amounts of available data for VLSM, SVR-LSM, SCCAN and DLM.}
    \label{fig:substrate_recover_varying_continuous_n_1}
\end{figure}

The results obtained for each method on all metrics are displayed in Figure \ref{fig:varying_n_continuous} and example substrate recoveries are presented in Figure \ref{fig:substrate_recover_varying_continuous_n_1}. Note that all methods perform better both quantitatively and qualitatively when the data has linear deficit labels versus binary labels. This is because each label includes more information about the target substrate which makes the regression task easier. Also methods that use permutation testing obtain more informative error rates when using continuous targets \cite{permutation_test_binary}. \newline

Inspection of the results in Figure \ref{fig:varying_n_continuous} shows that the multivariate methods behave as expected and outperform VLSM on all metrics and for all data regiments. Unlike with binary deficit labels, now SVR-LSM outperforms VLSM even on HD. As with the binary deficit setup, SCCAN outperforms SVR-LSM on HD, i.e., the false positives it predicts are closer to the ground truth substrate. Similarly, SVR-LSM outperfoms SCCAN on the Dice. Both models perform similarly on average surface distance. DLM outperforms all models on all data regiments and the relative improvements over baselines stay roughly the same, thus demonstrating that the framework established in \S \ref{vae_ldm_theory} performs well independently of the type of likelihood used to model the labels. \newline

\subsection{Noisy lesion-deficit relationship}

\begin{figure}[H]
\hspace*{-1.3cm}
    \centering
    \includegraphics[width=15cm]{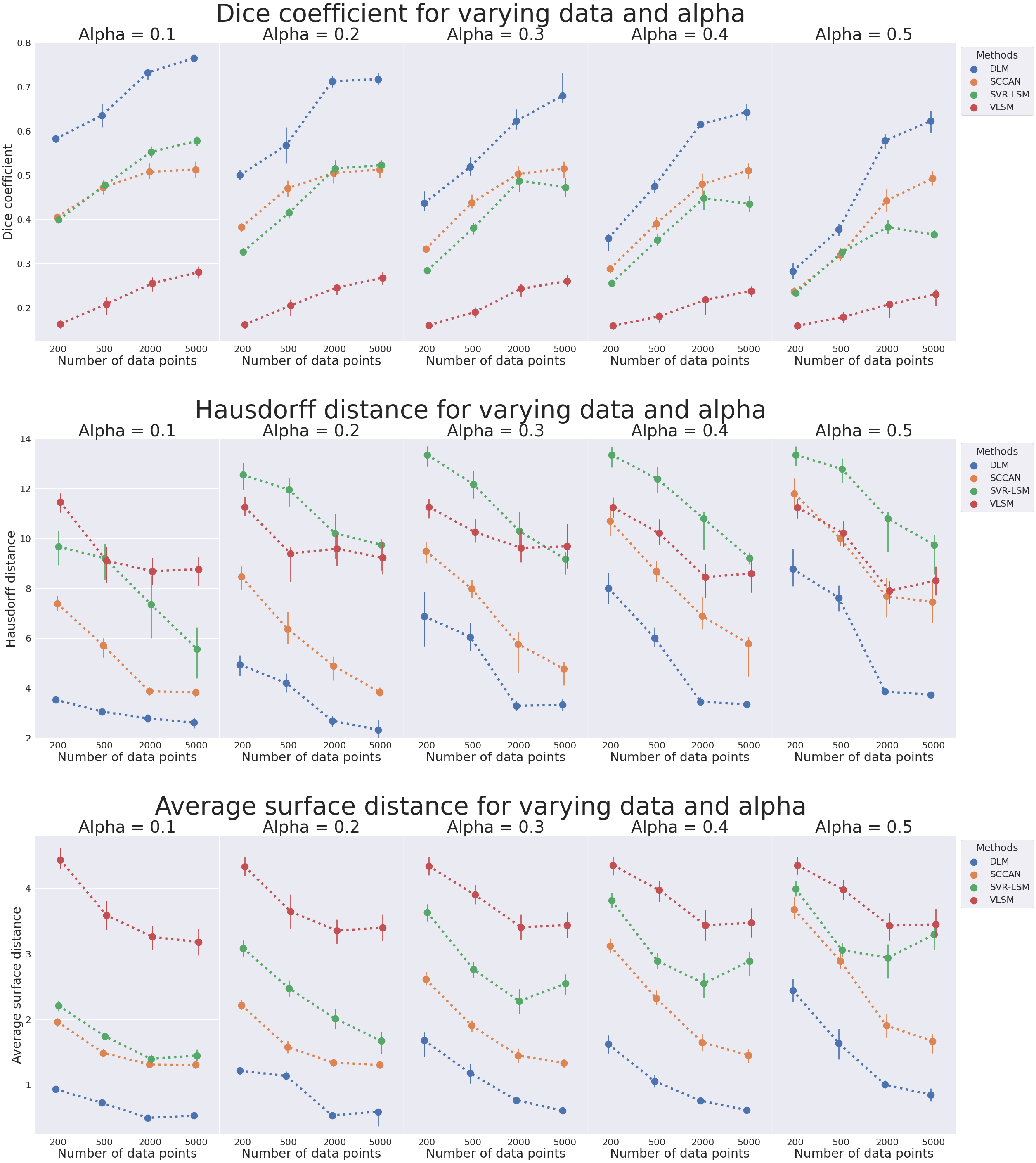}
    \caption[Model performances on the task of substrate recovery given continuous lesion deficit labels for varying amounts of data and varying amounts of noise]{Model performance on the task of substrate recovery given continuous lesion deficit labels for varying amounts of data and varying amounts of noise. The error bars correspond to 1 standard deviation. From left to right we vary the amount of noise, from top to bottom we vary the metric to be gauge. Dice (top), higher is better. Hausdorff distance (middle), lower is better. Average surface distance (bottom), lower is better.}
    \label{fig:varying_n_continuous_noise_1}
\end{figure}

\begin{figure}[H]
\hspace*{-1cm}
    \centering
    \includegraphics[width=15cm]{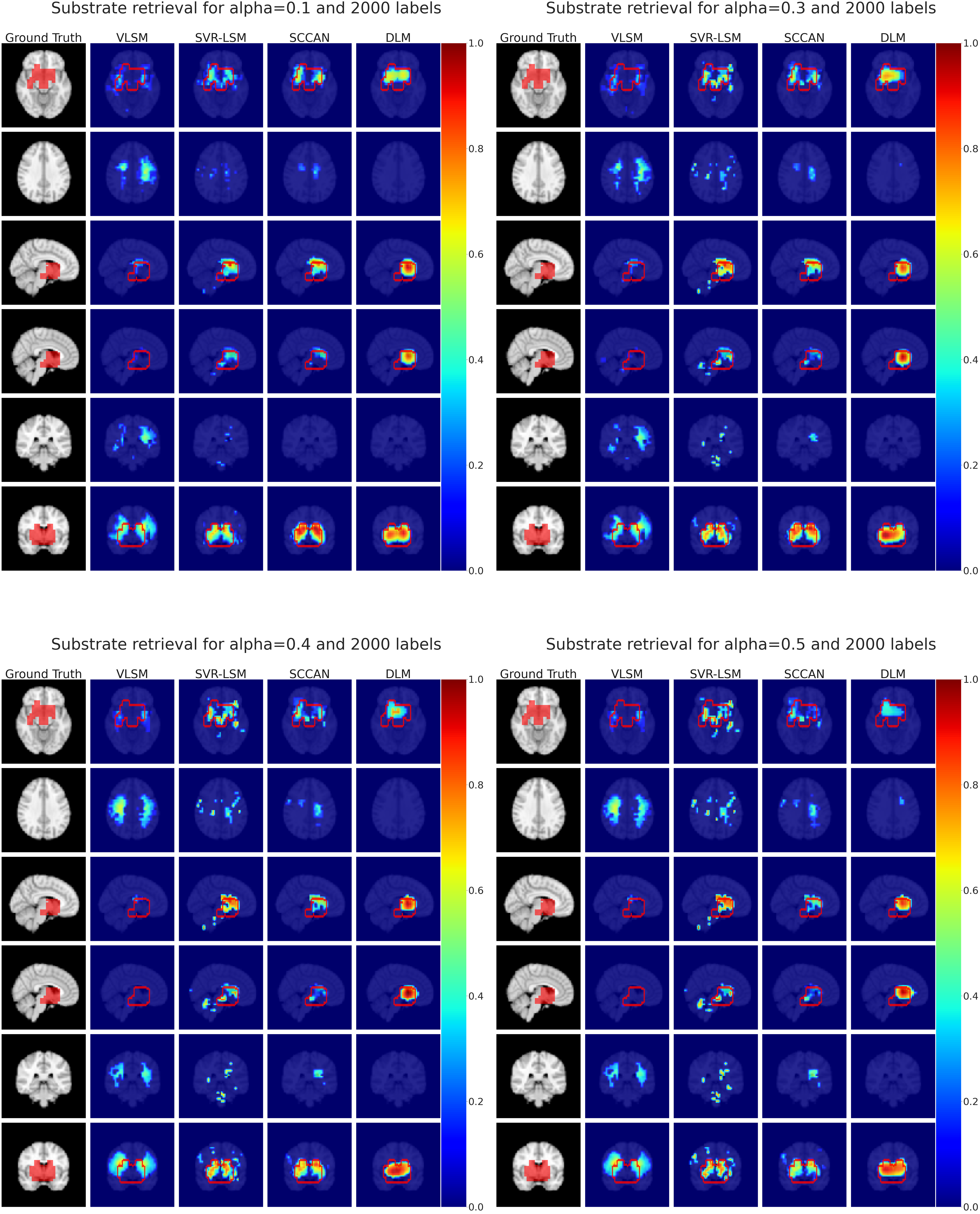}
    \caption[Example substrate recoveries for continuous lesion-deficit labels with varying amounts of uniform noise]{Example substrate recoveries for continuous lesion-deficit labels with varying amounts of uniform noise.}
    \label{fig:substrate_recover_varying_continuous_noise_11}
\end{figure}

We explore how model performance changes as we increase $\alpha$, i.e. the final label has less and less signal relating to the substrate. The model performances for varying amounts of noise and amounts of data are shown in Figure \ref{fig:varying_n_continuous_noise_1} and example substrate recoveries for various amounts of noise at $N=2000$ are presented in Figure \ref{fig:substrate_recover_varying_continuous_noise_11}. \newline

Our results for VLSM and SCCAN at varying levels of $\alpha$ seem to be roughly in line with the findings of \cite{pustina}, despite the nature of the data being slightly different. SVR-LSM suffers the largest deterioration with increase in $\alpha$, underperfoming VLSM on HD for $\alpha \geq 0.2$. Literature on SVR-LSM evaluation \cite{zhang_svr, svrlsm_eval} has been evaluated exclusively on clean labels and we hope our experiments raise awareness regarding its fragility to noise. To alleviate the issue one could reduce the amount of false positives the model produces by further increasing the amount of regularisation used. However, we found that leads to a marked increase in false negatives as well and therefore one enters a trade-off space between Dice score for Hausdorff distance. The results presented are for this reason obtained with the SVM hyperparameters that produced the best cross-validation performance as described in \S \ref{svm_setup}. Furthermore, SVM's ability to deal with noise depends on the kernel used. High-bias kernels such as linear or polynomial are not as affected by noise as a low-bias kernel like RBF, but high-bias kernels result in high numbers of false negatives. On a final note, SVMs can be made more robust to noisy labels \cite{robust_svm}, however, these approaches make inference impossible. \newline

Finally, we note that our method always outperforms all baselines for all values of $\alpha$, although here the deterioration of the model's performance as noise increases is slightly more pronounced than when using binary labels. Similarly to \S \ref{experiments_bin_noise} this set of experiments demonstrates that DLM despite having more parameters, overfits less to the training and validation data. Whereas for the binary case there was no explict modelling of variance, when training on continuous labels the model's resilience can also be justified by the fact that the log-likelihood $\log P_{\boldsymbol{\gamma}}(\mathbf{Y}|\mathbf{X},\mathbf{Z})$ is Gaussian and has its standard deviation directly modelled by the DLM which helps downweight labelled examples where the noise is high. \newline

\subsection{Non-linear lesion-deficit relationship}

We repeat the experiments from the previous section but now using a noisy non-linear lesion deficit relationship as described in \S\ref{non_linear_label_sect} and \ref{noise_explanation}. The most noticeable difference with this experimental setup is that now the deterioration of the performances of SVR-LSM as noise is increased are much more pronounced, further highlighting the model's vulnerability to overfitting. For every other model there is a slight deterioration of Dice when using non-linearities, which are qualitatively correlated with the amount of false negatives  seen in Figure \ref{fig:substrate_recover_varying_non_linear_noise_15}. This due to the lesion-deficit scores plateauing when a lesion surpasses a particular ratio of overlap (see Figure \ref{fig:non_linear_lesion_deficit}) and therefore some lesions will now be less informative with regards to the true positive voxels belonging to the substrate. \newline

\begin{figure}[H]
\hspace*{-1cm}
    \centering
    \includegraphics[width=14cm]{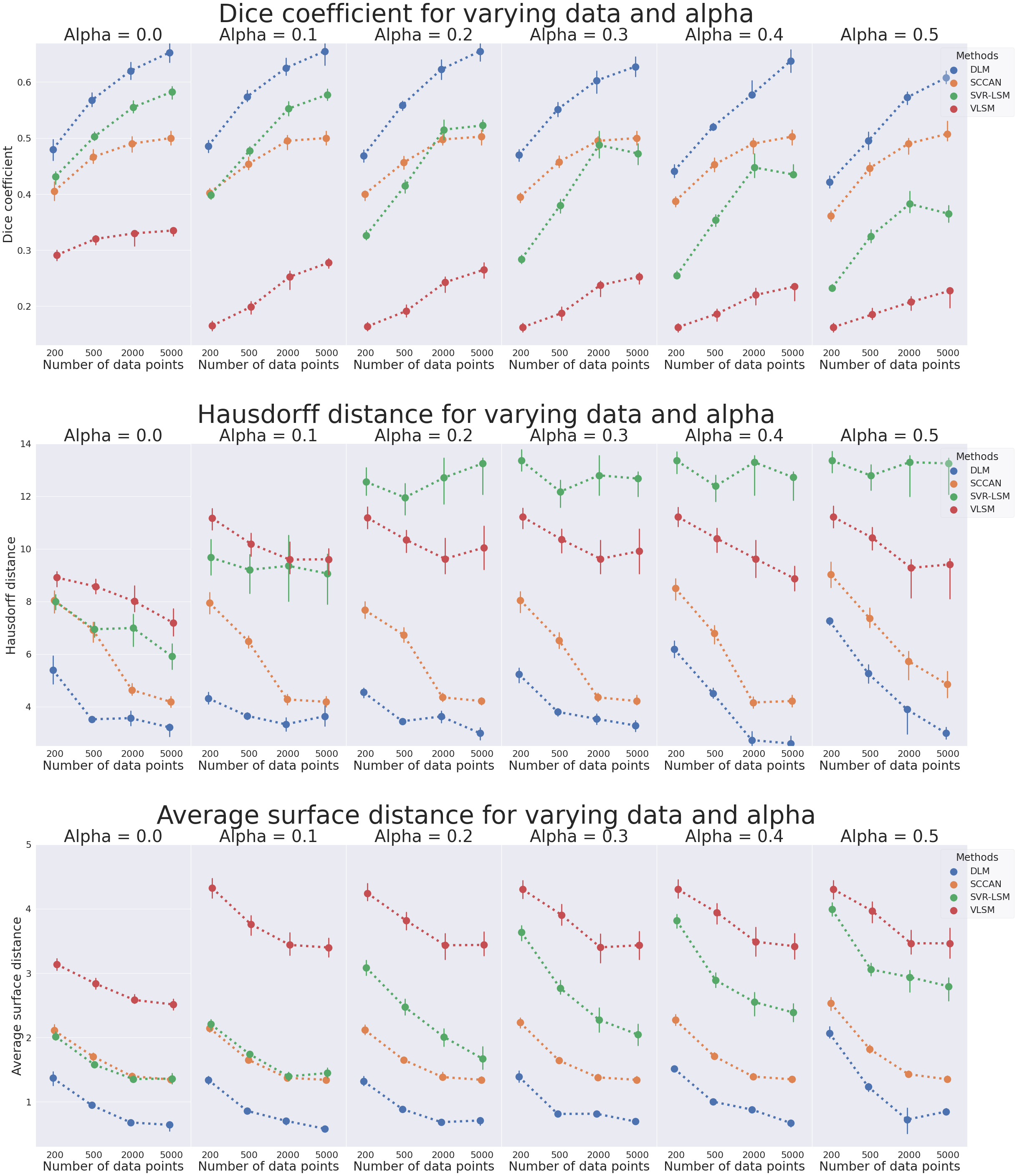}
    \caption[Model performances on the task of substrate recovery given non-linear continuous lesion deficit labels for varying amounts of noise]{Model performance on the task of substrate recovery given non-linear continuous lesion deficit labels for varying amounts of data and varying amounts of noise. The error bars correspond to 1 standard deviation. From left to right we vary the amount of noise, from top to bottom we vary the metric to be gauge. Dice (top), higher is better. Hausdorff distance (middle), lower is better. Average surface distance (bottom), lower is better.}
    \label{fig:varying_n_non_linear_noise_1}
\end{figure}

\begin{figure}[H]
\hspace*{-1cm}
    \centering
    \includegraphics[width=15cm]{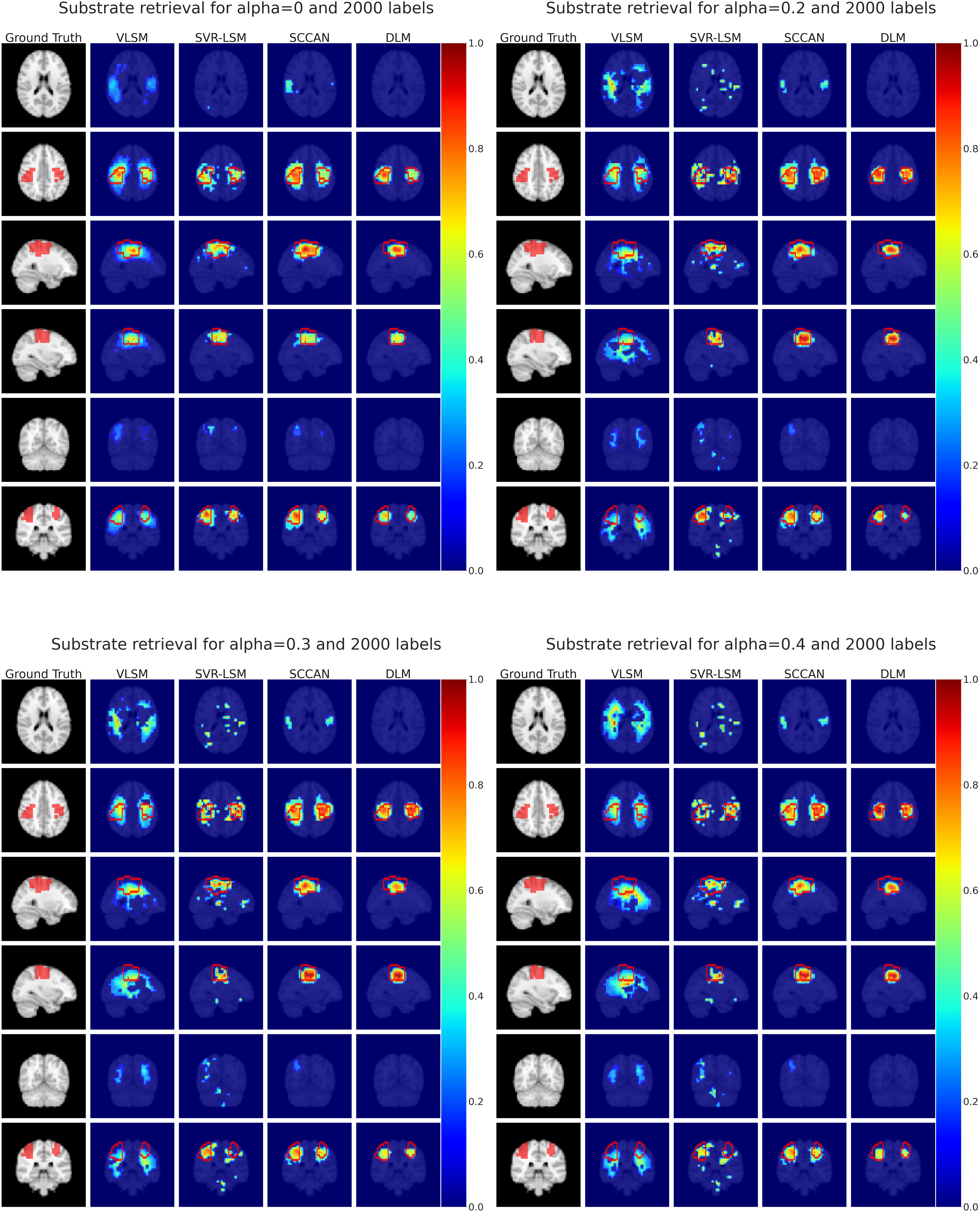}
    \caption[Example substrate recoveries for continuous non-linear lesion-deficit for varying amounts of uniform noise]{Example substrate recoveries for continuous non-linear lesion-deficit labels with varying amounts of uniform noise.}
    \label{fig:substrate_recover_varying_non_linear_noise_15}
\end{figure}

\subsection{Heterogeneous substrates}

Although noise simulates a type of heterogeneity in the patient population, it is important to consider heterogeneity at the level of the substrate topology. Similarly to \cite{sccan}, we introduce simulations where there are two possible functional substrates, $A$ and $B$, to recover given a patient population. When simulating behavioural scores, there is a 50\% chance the scores are calculated using substrate $A$ and a 50\% chance they are calculated with substrate $B$. We test multiple substrate combinations given the neural substrates described in \S \ref{functional_neural substrate_section} by simply pairing each neural substrate with every other neural substrate and then excluding combinations that do not allow for at least 10\% of the dataset to touch that resulting substrate. Due to the increased complexity of the substrates we now only evaluate models using at least $N=1000$ data points to guarantee sufficient lesion sampling of the substrate. \newline

\begin{figure}[H]
\hspace*{-1.5cm}
    \centering
    \includegraphics[width=12cm]{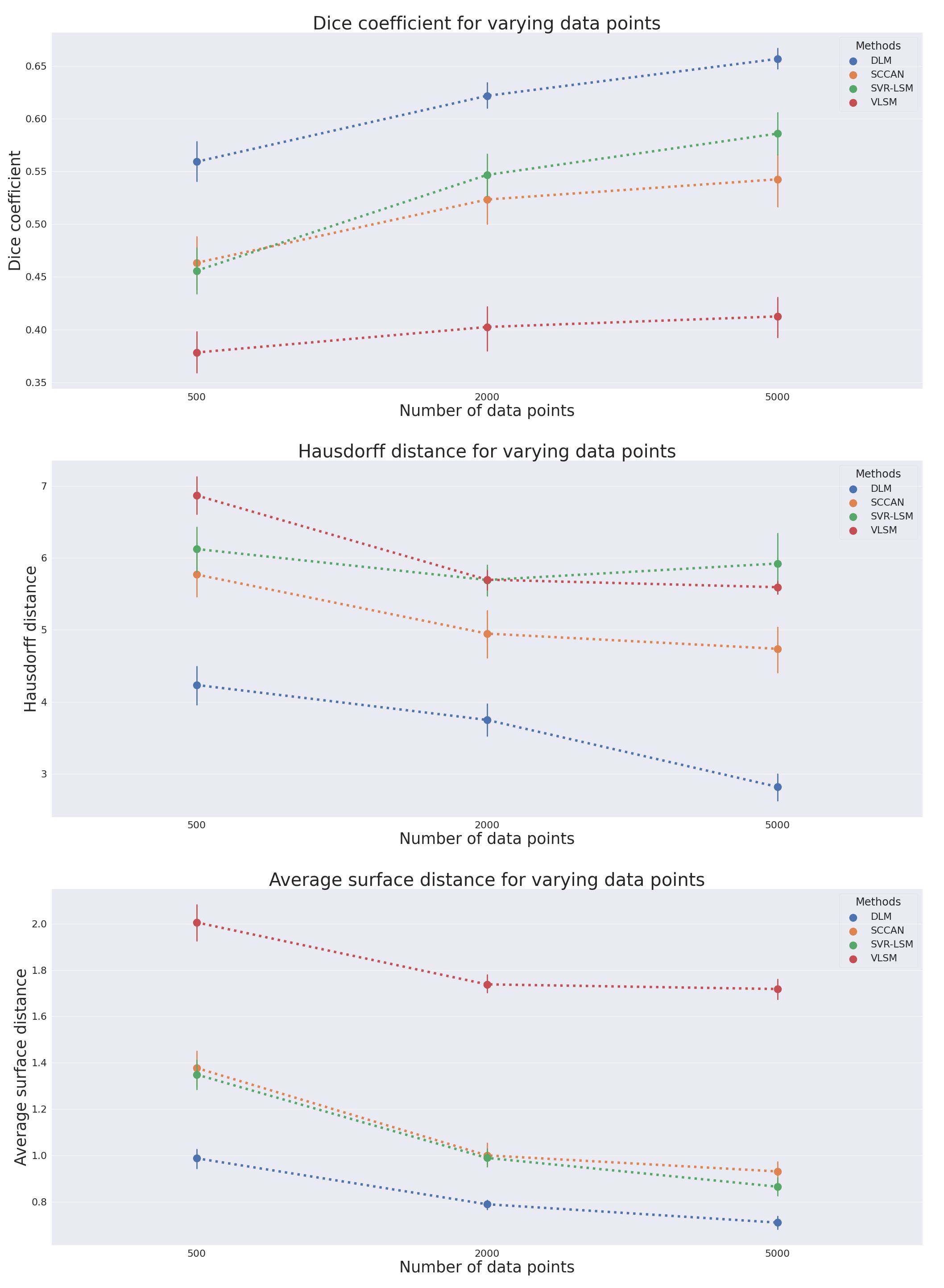}
    \caption[Model performance on the task of substrate recovery given continuous lesion deficit labels and heterogeneous ground truth substrates]{Model performance on the task of substrate recovery given continuous lesion deficit labels and heterogeneous ground truth substrates for varying amounts of data. The error bars correspond to 1 standard deviation. Dice (top), higher is better. Hausdorff distance (middle), lower is better. Average surface distance (bottom), lower is better.}
    \label{fig:varying_n_linear_hetero}
\end{figure}

\begin{figure}[H]
\hspace*{-0.5cm}
    \centering
    \includegraphics[width=15cm]{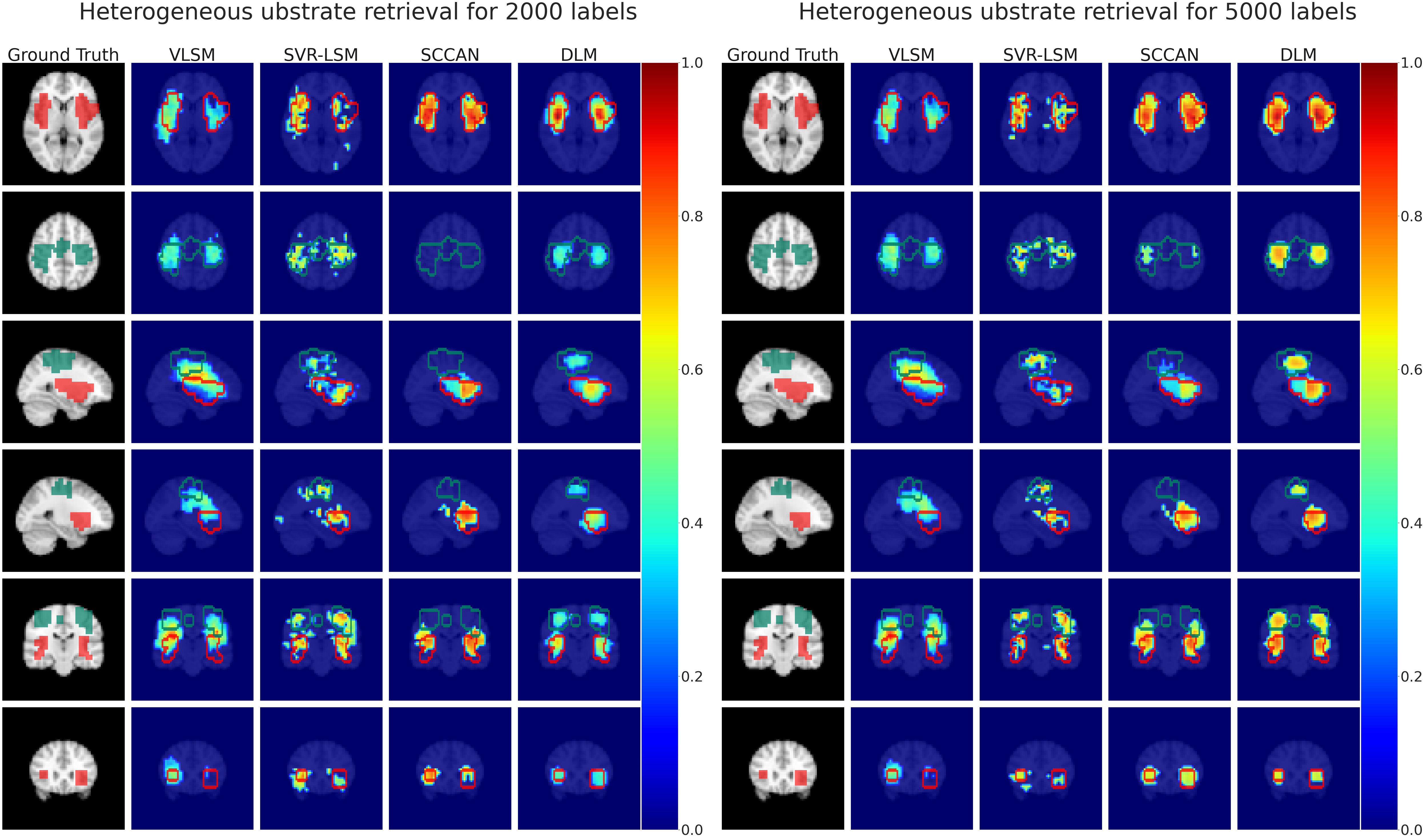}
    \caption[Example substrate recoveries for a heterogeneous ground truth and continuous labels]{Example substrate recoveries for a heterogeneous ground truth and continuous labels}
    \label{fig:substrate_recover_varying_hetero_linear_11}
\end{figure}

As expected and shown in Figures \ref{fig:varying_n_linear_hetero} and \ref{fig:substrate_recover_varying_hetero_linear_11} using heterogeneous substrates results in a decrease in performance on all models when compared to homogeneous substrates, regardless of the amount of labelled data used. The most significant decreases are seen with both VLSM and SCCAN. These two models, tend to predict the midway point between the two possible substrates as is illustrated in Figure \ref{fig:substrate_recover_varying_hetero_linear_11}. This results in poor Dice and ASD performances due to the high number of false negatives. Despite the unimodality of their predictions both VLSM and SCCAN still have better HD performances than SVR-LSM, in a pattern that is consistent with the results of the previous experiments, where it has been shown that SVR-LSM produces distant false positives. \newline

The DLM model outperforms all models across all data regiments, but even at $N=5000$ it still produces a high number of false negatives. If functional substrates in clinical scenarios are indeed heterogeneous, then this further highlights the need for large data regiments to accurately perform lesion-deficit mapping. 

\subsection{Spatial bias of model predictions}
\label{spatial_bias}

Following \cite{yee_parashkev}, we illustrate the extent of mislocalization resulting from interactions between the lesion distribution and the underlying substrate. To achieve this, we simulate ground truth substrates that are a single voxel, and for all the lesions in our dataset we label them `affected', $y=1$, if they intersect with that single voxel or `unaffected', $y=0$, otherwise. Operating at 2mm resolution, we choose slices in the sagittal (8100 voxels, $x=50$,), coronal (9720 voxels, $y=28$) and axial (9720 voxels, $z=42$) orientations and for every other voxel in these slices, we create a lesion-deficit experiment with our entire lesion dataset where said voxel is critical to a hypothetical deficit. In order to not be biased by the nature of the ischaemic stroke distribution, we only use voxel locations hit at least four times in the data set. Due to the nature of this simulation the task is binary, however, one could simulate continuous targets by using larger ground truth substrates, at the cost of a deterioration in the resolution of the mislocalization results. \newline

\begin{figure}[H]
\hspace*{-1cm}
    \centering
    \includegraphics[width=15cm]{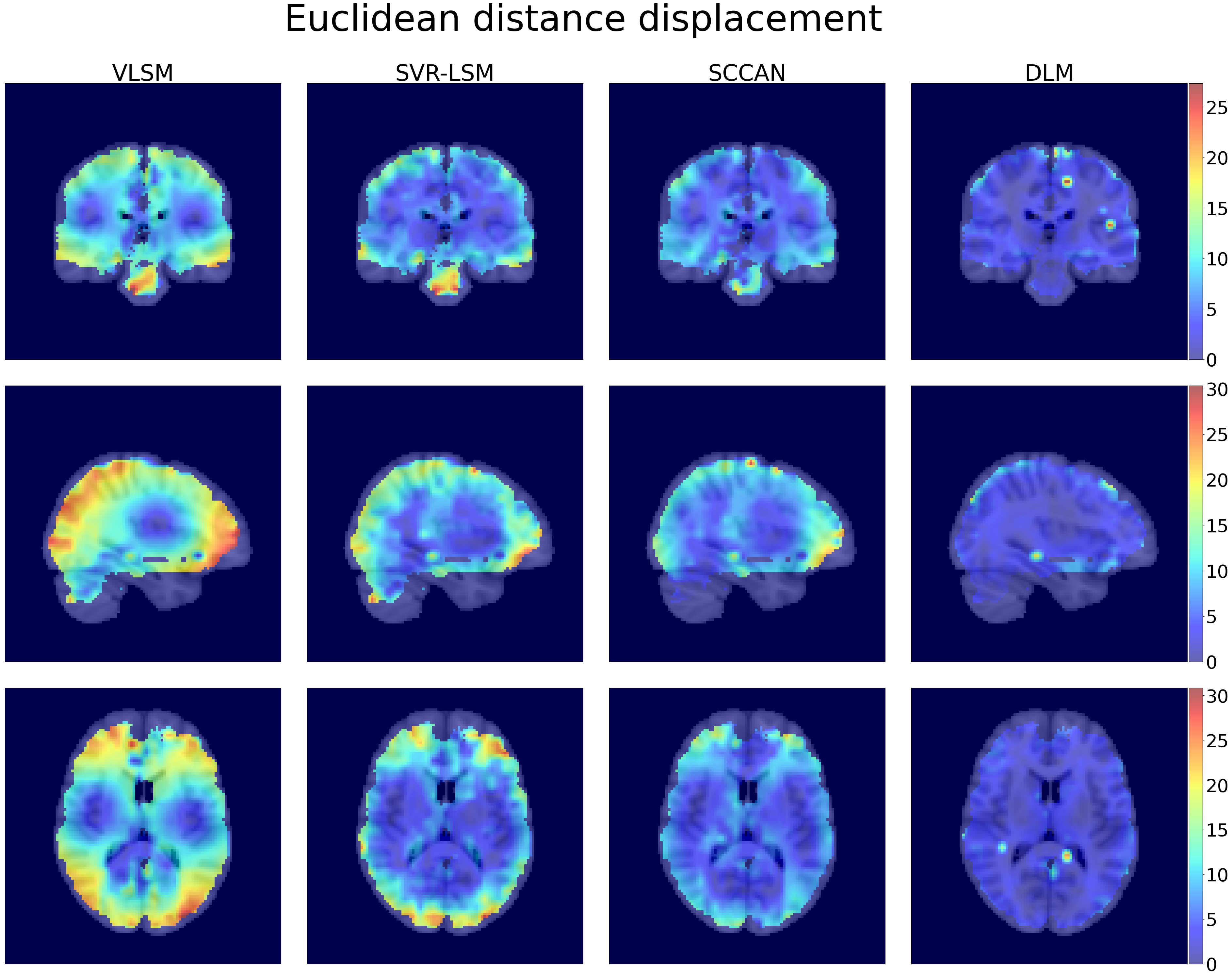}
    \caption[The magnitude of the inferential displacement for each of the models]{The magnitude of the displacement for each of the models}
    \label{fig:magnitude_displacement}
\end{figure}

\begin{figure}[H]
\hspace*{-1cm}
    \centering
    \includegraphics[width=15cm]{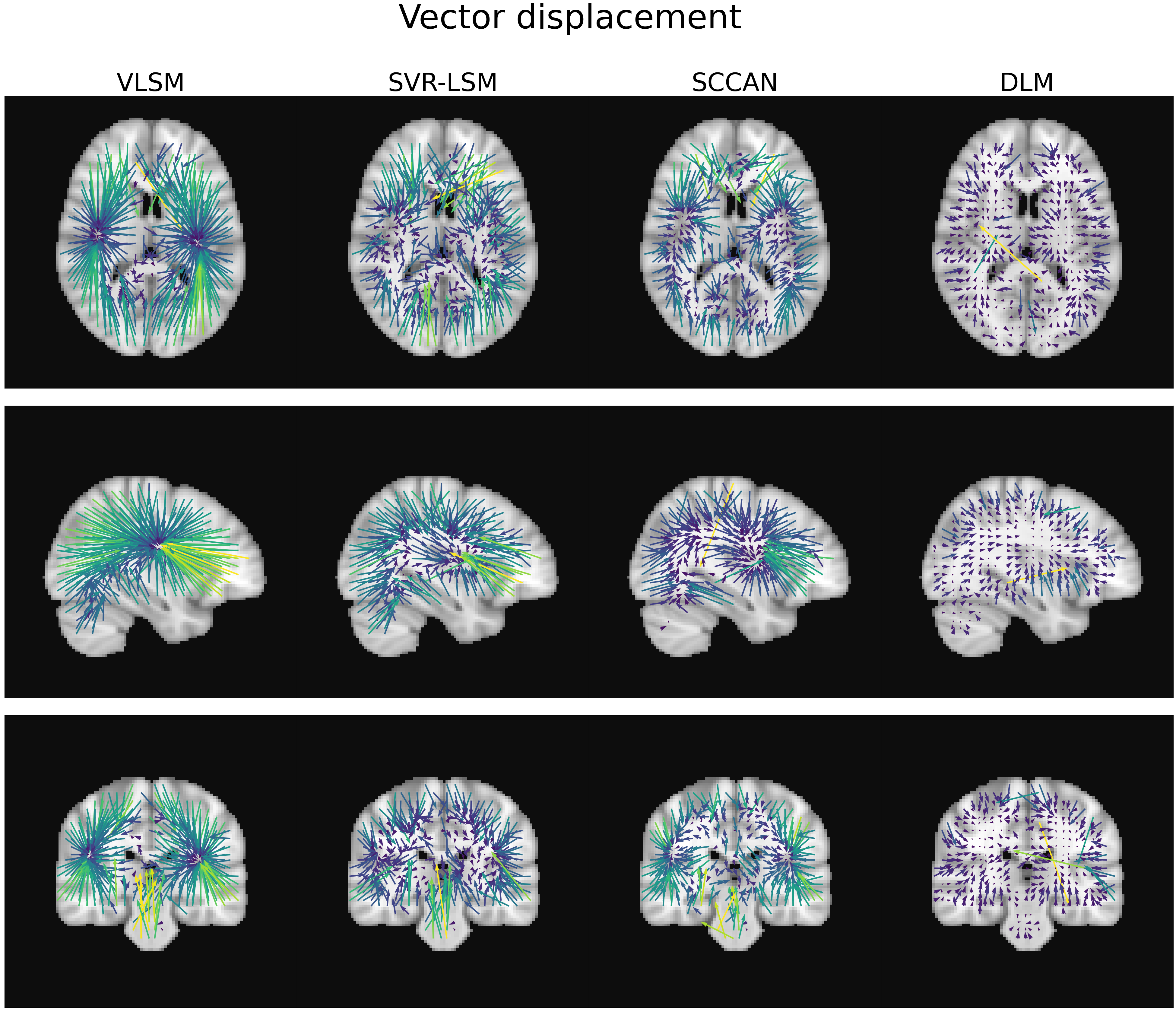}
    \caption[The directionality of the inferential displacement for each of the models]{The direction of the displacement for each of the models}
    \label{fig:direction_displacement}
\end{figure}

Given the set of experiments, we train every single baseline model to obtain a predicted cluster of voxels. Given this cluster has finite set of points $\{p_1, \ldots, p_k\}$ in $\mathbb{N}^2$ we calculate its centroid using a mean of those points, thus obtaining a centroid which minimizes the sum of squared Euclidean distances between itself and each point in the set \cite{analytic_geometry}. Given the predicted centroid for each model, we then calculated its displacement from the label-defining voxel, producing a two-dimensional vector that represents the error introduced by each of the evaluated methods. The magnitude of these errors is illustrated in Figure \ref{fig:magnitude_displacement} and Table \ref{displacements_table}. The VLSM values on the table roughly match the values presented in \cite{svrlsm_eval}. The directionality of the displacement is illustrated in Figure \ref{fig:direction_displacement}. \newline

Inspection of the table and both figures shows that the spatial displacement reflects the findings of previous sections. VLSM, unsurprisingly, has the highest displacement. SVR-LSM and SCCAN perform roughly equally, with SCCAN having on average 1mm less displacement. We hypothesise this is related to SVR-LSM's consistently higher Hausdorff distances which are symptomatic of distant false positive predictions. \newline

Our model, DLM, has the smallest displacements, which is expected given its superiority in performance for Dice, HD and ASD across the board. However, we note that despite its consistently low average error DLM still produces some outliers with high displacement. They are infrequent and hence do not cause an elevated standard deviation displacement \ref{fig:magnitude_displacement}. These outliers plausibly arise as a result of training DLM only once per voxel ground truth owing to the computational burden of this specific validation. In previous experiments, the model is run 3 times on different permutation splits of the data. Neural networks are stochastic models and therefore can converge to slightly different solutions even with the same initial set of hyperparameters. In real-world practice where such limitations do not arise, we would be expected to train the model multiple times, taking the average of the inferred map.
\begin{table}[h]
\centering
\begin{tabular}{ll} \toprule
 Method & Displacement (mm) \\ \midrule
 VLSM &  8.5 $\pm$ 5.6 \\
 SVR-LSM &  5.5 $\pm$ 3.7 \\
 SCCAN & 4.5 $\pm$ 3.0 \\
 \textbf{DLM} &  \textbf{2.0} $\pm$ \textbf{1.3} \\
 \bottomrule
\end{tabular}
\caption{Average and standard deviation of displacement vectors for every baseline model}\label{displacements_table}
\end{table}

\section{Discussion}

The extraordinary complexity of the human brain makes inferences to its functional organisation with correlative data alone fundamentally insecure. Yet the choice of methods capable of establishing necessity here is very narrow: indeed, lesion-deficit mapping is arguably the only one with comprehensive anatomical coverage. The fidelity of functional anatomical inferences drawn from lesions ought, then, to be of the highest importance, for we rely on them to confirm or infirm what other methods may suggest but not determine conclusively. \newline

So crucial a role is supported by surprisingly little attention to the appropriate inferential framework. Mass-univariate methods remain the most widely used, despite incontrovertible empirical evidence of their unsuitability and---above all---clear violations of their foundational assumptions. Where a lesion spans more than a single unit of anatomical analysis, no hypothesis of neural dependence is testable at a single locus in isolation from all others, for the deficit is caused by \textit{the lesion as a whole.} Only multivariate approaches, inevitably within the framework of model comparison rather than hypothesis testing, provide a principled basis for inference here.\newline

A major obstacle to methodological development is the absence of a ground truth. The validity of inferences about a putative substrate rests heavily on the foundational assumptions of the model. A model perfectly predictive of a deficit may identify none of the critical areas; conversely, a poorly predictive model may be topologically faithful. Until a consensus emerges on the acceptable level of logical hygiene in the field, this leaves semi-synthetic numerical simulations as the only means of comparing the fidelity of alternative inferential approaches. \newline

Since no single simulation can be assumed to \textit{replicate} reality, any rigorous numerical validation must explore a range of parameters wide enough plausibly to \textit{enclose} it. And since the parameter space is extraordinarily large, here we apply the broadest empirical priors, based on the largest scale data available, and constrain our simulations primarily by feasibility. In the largest and most comprehensive analysis of its kind, we evaluate 5500 acute ischaemic lesions from 4788 unique patients with 20 combinations of 16 different hypothetical neural substrates drawn from hierarchical clustering of large-scale meta-analytic functional imaging data, across Z lesion-deficit simulations. We quantify the impact on fidelity of substrate anatomy, sample size, and label noise, in the context of binary, linear, non-linear, homogeneous, and heterogeneous lesion-deficit relationships, establishing a candidate standard benchmark framework for evaluating lesion-deficit models of any kind. \newline

Whereas semi-synthetic lesion-deficit simulations are helpful in \textit{validating} a model, they \textit{cannot also} be used to \textit{design or train} it, for obvious reasons of circularity. The approach we develop here, DLM, is therefore directed by the fundamental nature of the task and the plausible underlying generative process. As we outline in the introduction, lesion-deficit mapping requires a sufficiently expressive multivariate model capable of capturing the \textit{joint} distribution of lesions and deficits in anatomical terms. This implies a generative model of volume data; given the pressure on maximising expressivity, most naturally a deep generative one. \newline

DLM builds on recent advances in volumetric deep generative models \cite{diffeo_gan, pixelcnn}, initiating the study of applying such models to the task of three dimensional topological inference. The theoretical framework we present in \S \ref{vae_ldm_theory}, models the joint distribution of lesions and associated deficits with an appropriately crafted VAE \cite{vae, vae_baur, vae_text}, but can be implemented with any latent variable model. We provide a self-contained, compact implementation of this framework in \S \ref{network_architecture} that can be applied to any lesion-deficit problem with minimal user intervention. \newline

Comprehensively evaluated on our semi-synthetic framework, DLM shows superior fidelity over VLSM across all lesion-deficit scenarios by a substantial margin, in reflection of the now familiar shortcomings of mass-univariate techniques \cite{univariate_shortcomings_1, univariate_shortcomings_2, pustina, zhang_svr, yee_parashkev}. The superiority extends to the two currently most popular multivariate methods, SCCAN \cite{pustina} and SVR-LSM \cite{zhang_svr}, again across all modelled lesion-deficit scenarios. Here we consider nine points arising in the interpretation of these results.



First, DLM's comparative success ought to be unsurprising given that it is the first method to address each of the component demands of the task, especially the capacity to capture distributed spatial lesion-deficit relations within a sufficiently expressive generative framework. Neither mass-univariate nor discriminative multivariate methods are suited to the kind of question lesion-deficit inference poses; their weakness here is not a fault of design but a consequence of misapplication. Nor is it plausibly attributable to our specific implementation of DLM: the decisive characteristics of the model are theoretically achievable with other suitable volumetric deep generative architectures. Indeed, in parallel with providing a robustly validated tool, ready for application to real-world lesion-deficit problems, we wish to stimulate the wider development of this approach to spatial inference, within the present research domain and outside it. \newline

Second, DLM's performance metrics must be interpreted comparatively, by reference to competing models deployed on the same task, in the same circumstances. We cannot compare the absolute Dice scores here with those judged acceptable for other imaging tasks such as lesion segmentation, where a ground truth is available and the demands are very different. The maximum achievable fidelity of substrate recovery is here limited by the characteristics of the lesions used to perform the inference: they can be conceived of as a lossy, distorting filter no model could perfectly correct. Though our analysis casts some light on the theoretical resolvability of neural substrates with acute ischaemic lesions, its primary objective is to determine which one of a set of rival inferential methods a lesion-deficit mapper is best advised to employ in any given study. The clear answer, for now, is DLM, but we hope to stimulate the development and competitive evaluation of other architectures. \newline          

Third, we should be clear that none of the architectures evaluated here can be driven to overfit \textit{on the key task of spatial inference}, even if they may well overfit on deficit prediction, for they are not exposed to substrate maps at any point during model development or training. Indeed, a tendency to overfit on the prediction would result in impaired inferential performance resulting from erroneous weighting of lesion features collaterally associated with the outcome. That DLM has a larger number of parameters than the other multivariate models does not therefore materially increase the risk of inflated performance on that account. Nor does the use of a validation set to determine the binarization threshold incur a similar risk, with DLM or SSCAN, for again the criterion is predictive performance, not spatial retrieval. There is, of course, always the possibility of an accidental overfit, here minimized by extensive evaluation across the widest range of lesion-deficit scenarios ever attempted.  

Fourth, it may be surprising to see a richly parameterised model outperform both mass-univariate and multivariate methods across all data regimes, including small data scales. Three architectural characteristics arguably explain this: 
\begin{itemize}
    \item Our generative model incorporates a `meta-learning' mechanism \cite{meta_learning, meta_vaes}. Though the VAE has many parameters, the log-likelihood $\log P_{\boldsymbol{\gamma}}(\mathbf{Y}|\mathbf{X},\mathbf{Z})$ is predicted exclusively from the latent substrate $\hat{\textbf{m}}$, which has only as many parameters as there are input features. This acts as a powerful regularizer. 
    \item The variational nature of the model means that the data passed through the decoders is augmented with noise. Binary lesions and their associated deficits cannot be augmented in the traditional ways used in computer vision, since any type of augmentation risks changing the underlying semantics. By adding noise to the feature space learnt by the model, however, a VAE framework enables effective data augmentation without changing the semantic meaning of the original input. The robustness to overfitting of variational Bayesian methods is well known \cite{variational_robustness}.
    \item In the context of noisy labels, sampling the latent variables $\mathbf{z}$ for the posterior from a learnt multi-dimensional mean, $\boldsymbol{\mu}$ and standard deviation, $\boldsymbol{\sigma}$ allows noise in the labels used to model the log-likelihood $\log P_{\boldsymbol{\gamma}}(\mathbf{Y}|\mathbf{X},\mathbf{Z})$ to be accounted for via $\boldsymbol{\sigma}$
\end{itemize}

Fifth, DLM is computationally expensive in proportion to the size and image resolution of the data. It is nonetheless operable on consumer-grade GPUs, and may even be run on the CPU over timescales acceptable in the context of lesion-deficit research studies. The computational demands of a methodological validation study involving thousands of numerical experiments are, of course, much greater. For this reason, here we used the same DLM hyperparameters in every experiment. As discussed in \S \ref{spatial_bias}, this means performance would likely have been underestimated. We did not solicit the additional computational resource needed to perform hyperparameter tuning because performance was decisively superior to all other baselines (which were hyper-parameter tuned as applicable).\newline

Sixth, the main user configurable parameter of DLM is the image resolution at which the lesion-deficit association is estimated. DLM's fully convolutional architecture confers resilience to changes in resolution, but it is natural to select a scale appropriate to the density and variability of lesion sampling. Equally, DLM's spatial decomposition of a lesion may be constrained by a region-of-interest parcellation of the whole brain or a target territory, but that inevitably introduces a spatial bias that may or may not correspond with reality. Where the task is true spatial inference---not selecting between regions spatially delineated \textit{a priori}---it seems to us appropriate to limit the use of strong spatial priors, for otherwise the framing of the question may unacceptably limit the range of possible answers. 

Seventh, in common with widespread practice in the field, the inferred map does not provide a weighting of the relative contribution of \textit{components} of the substrate, or of their (potentially complex) relations. The lesion-deficit function \textit{within the critical substrate} is assumed to be comparatively simple, in line with current models of the macroscopic organisation of the brain. Such inference arguably requires modelling of the substrate as a graph \cite{cipolotti2023graph}, currently computationally feasible only at comparatively coarse resolutions. DLM may, however, be naturally extended to graphs through variational graph autoencoder \cite{kipf2016variational} architectures recently shown to be adaptable to graphs of substantial size \cite{salha2021fastgae}. 

Eighth, again in line with established practice, the lesion at the input and the inferred substrate at the output are binarized images. A lesion may be represented as an anatomical continuity of neural dysfunction, either because the pathological process is graded, or because its effects are variably propagated by (say) disconnected white matter tracts. Equally, neural substrates may exhibit a continuous anatomical organisation, following local or global gradients. Continuous lesion representations such as disconnectomes can be modelled within the existing DLM framework without any modification, for their modelled interaction with the substrate remains unaltered except for corresponding voxel-wise weighting (cf \S\ref{vae_ldm_theory}). Modelling continuous neural substrates also requires no modification beyond omitting the final thresholding step. Naturally, it would be difficult to distinguish between local variations in the inferred substrate arising from inter-subject variability vs different magnitudes of neural dependence.          


Ninth, DLM may be naturally reformulated in semi-supervised form to draw value from unlabelled lesion data \cite{semisupervised_vae}. We can use such data to optimise the model by marginalising out the label:

\begin{equation}
\begin{split}
    P_{\boldsymbol{\theta}}(\mathbf{X}) &= \int_{Y} \int_{X} P_{\boldsymbol{\theta}}(\mathbf{X}, \mathbf{Y}| \mathbf{Z}) P(\mathbf{Z}) d\mathbf{Y} d\mathbf{Z} \\
   &= \mathbb{E}_{(\mathbf{y}, \mathbf{z})\sim Q_\phi(\mathbf{Y}, \mathbf{Z} | \mathbf{X})} \frac{P_{\boldsymbol{\theta}}(\mathbf{X}, \mathbf{Y}| \mathbf{Z}) P(\mathbf{Z})}{Q_\phi(\mathbf{Y}, \mathbf{Z} | \mathbf{X})}.
\end{split}
\end{equation}

Using properties of conditional probability we write the posterior as:

\begin{equation}
    Q_\phi(\mathbf{Y}, \mathbf{Z}| \mathbf{X}) =  Q_\phi(\mathbf{Y}|\mathbf{Z}, \mathbf{X}) Q_\phi(\mathbf{Z} | \mathbf{X}).
\end{equation}

Through the use of logarithms and Jensen's inequality we get the following variational lower bound:

\begin{equation*}
\mathbb{E}_{\mathbf{y}\sim Q_\phi(\mathbf{Y}|\mathbf{Z}, \mathbf{X})} \mathbb{E}_{\mathbf{z}\sim Q_\phi(\mathbf{Z} | \mathbf{X})} \big[\log P_\theta(\mathbf{X},\mathbf{Y}| \mathbf{Z})
\end{equation*}
\begin{equation}
+ \log P(\mathbf{Z}) - \log Q_\phi(\mathbf{Y}|\mathbf{Z}, \mathbf{X}) -\log Q_\phi(\mathbf{Z} | \mathbf{X}) \big].
\end{equation}

Using the factorisation of the log likelihood introduced in \S \ref{vae_ldm_theory}, we can then write a computable expression of the VLB in terms of KL divergences and a differential entropy term as follows:

\begin{equation*}
H(Q_\phi(\mathbf{Y}|\mathbf{X}, \mathbf{Z})) + \mathbb{E}_{\mathbf{y} \sim Q_\phi(\mathbf{Y}|\mathbf{Z}, \mathbf{X})}
    \big[ \mathbb{E}_{\mathbf{z}\sim Q_{\boldsymbol{\phi}}} \big[\log P_{\boldsymbol{\gamma}}(\mathbf{Y}|\mathbf{X},\mathbf{Z})
\end{equation*}
\begin{equation}
     + \log P_{\boldsymbol{\theta}}(\mathbf{X}|\mathbf{Z})\big] - D_{KL}(Q_{\boldsymbol{\phi}}(\mathbf{Z}|\mathbf{X}) || P(\mathbf{Z}))
    \big]. 
\end{equation}

\section*{Acknowledgments and Funding}

This research has been conducted using the UK Biobank Resource under Application Number 16273. This work is supported by the EPSRC-funded UCL CDT in Medical Imaging (EP/L016478/1), the Wellcome Trust (213038) and NIHR UCLH Biomedical Research Centre.

\bibliographystyle{unsrt}  
\bibliography{references}

\end{document}